\pdfoutput=1

\documentclass[11pt]{article}
\usepackage[dvipsnames]{xcolor}
\usepackage[preprint]{acl}

\usepackage{times}
\usepackage{latexsym}
\usepackage[T1]{fontenc}
\usepackage[utf8]{inputenc}
\usepackage{microtype}
\usepackage{inconsolata}
\usepackage{graphicx}
\usepackage{subcaption}
\usepackage{booktabs}
\usepackage{multirow}

\usepackage{algorithm}
\usepackage{algorithmic}

\usepackage{enumitem}
\usepackage{bbm}
\usepackage{amsthm}
\usepackage{amsmath}
\usepackage{amsfonts}
\usepackage{amssymb}

\newtheorem{definition}{Definition}

\usepackage{inconsolata}
\usepackage{listings}
\lstdefinestyle{correct}{
    basicstyle=\ttfamily,
    keywordstyle=\color{blue}, 
    morekeywords={SELECT, FROM, WHERE, AND, OR, UNION, INTERSECT, JOIN, ON, MIN, INNER, AS, MAX, AVG, BY, LIMIT, COUNT, GROUP, ORDER, ASC, DESC},
    columns=fullflexible, 
    keepspaces=true,
    backgroundcolor=\color{ForestGreen!15},
}
\lstdefinestyle{incorrect}{
    basicstyle=\ttfamily,
    keywordstyle=\color{blue}, 
    morekeywords={SELECT, FROM, WHERE, AND, OR, UNION, INTERSECT, JOIN, ON, MIN, INNER, AS, MAX, AVG, BY, LIMIT, COUNT, GROUP, ORDER, ASC, DESC},
    columns=fullflexible, 
    keepspaces=true,
    backgroundcolor=\color{red!15},
}

\usepackage{microtype}
\usepackage[noorphans,vskip=0ex]{quoting}
\usepackage{titlesec}

\expandafter\def\expandafter\normalsize\expandafter{%
    \normalsize%
    \setlength\abovedisplayskip{2pt}%
    \setlength\belowdisplayskip{2pt}%
    \setlength\abovedisplayshortskip{2pt}%
    \setlength\belowdisplayshortskip{2pt}%
}

\newcommand*\samethanks[1][\value{footnote}]{\footnotemark[#1]}

\title{Calibrating LLMs for Text-to-SQL Parsing by Leveraging Sub-clause Frequencies}

\author{%
  Terrance Liu\thanks{Work done during an internship at Bloomberg} \\
  Carnegie Mellon University\\
  \And
  Shuyi Wang \\
  Bloomberg \\
  \And
  Daniel Preo\c{t}iuc-Pietro \\
  Bloomberg \\
  \AND
  Yash Chandarana\thanks{equal contribution} \\
  Bloomberg \\
  \And
  Chirag Gupta\samethanks \\
  Bloomberg \\
}

\begin{document}

\maketitle

\begin{abstract}

While large language models (LLMs) achieve strong performance on text-to-SQL parsing, they sometimes exhibit unexpected failures in which they are confidently incorrect. Building trustworthy text-to-SQL systems thus requires eliciting reliable uncertainty measures from the LLM. In this paper, we study the problem of providing a calibrated confidence score that conveys the likelihood of an output query being correct. Our work is the first to establish a benchmark for \textit{post-hoc} calibration of LLM-based text-to-SQL parsing. In particular, we show that Platt scaling, a canonical method for calibration, provides substantial improvements over directly using raw model output probabilities as confidence scores. Furthermore, we propose a method for text-to-SQL calibration that leverages the structured nature of SQL queries to provide more granular signals of correctness, named ``sub-clause frequency" (SCF) scores. Using multivariate Platt scaling (MPS), our extension of the canonical Platt scaling technique, we combine individual SCF scores into an overall accurate and calibrated score. Empirical evaluation on two popular text-to-SQL datasets shows that our approach of combining MPS and SCF yields further improvements in calibration and the related task of error detection over traditional Platt scaling. 

\end{abstract}

\section{Introduction}

\begin{figure}[t!]
    \centering
    \includegraphics[width=0.9\columnwidth]{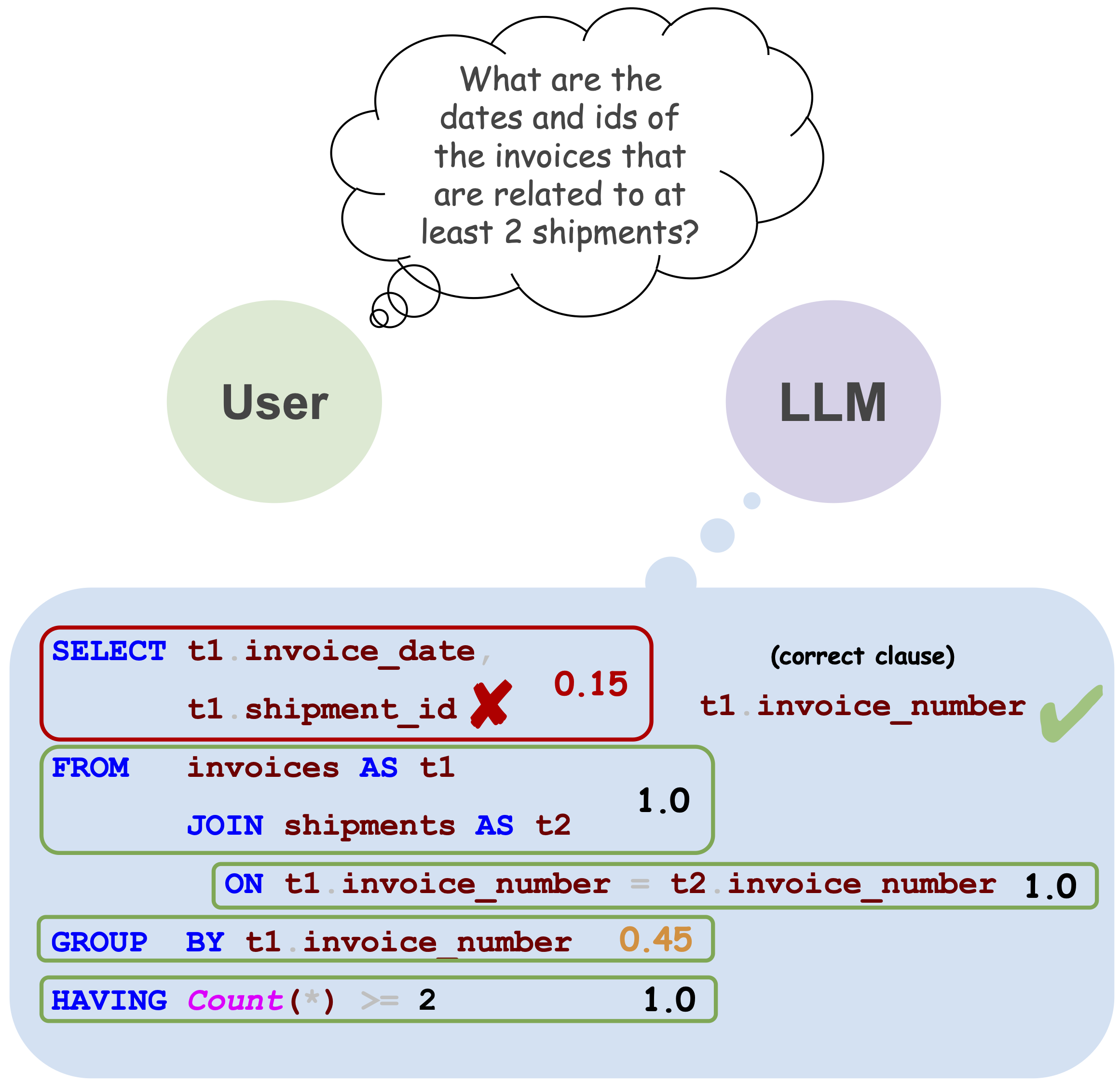}
    \caption{
        An example question from the SPIDER dataset and an output produced by \textsc{T5 3B}. Each box in the SQL output corresponds to a separate sub-clause, identified using a standard parser. The boundaries of the boxes in \colorbox{ForestGreen!15}{green} and \colorbox{red!15}{red} denote sub-clauses that are correct and incorrect (respectively) with respect to a ground truth query. In each box, we have shown a sub-clause frequency (SCF) score computed by our method. The SCF corresponds to the proportion of outputs that contain each sub-clause among additional outputs sampled via nucleus sampling and beam search (Figure \ref{fig:parsing_example} illustrates this further). SCF scores contain a granular signal for correctness. For example, here, the incorrect \texttt{SELECT} clause (which contains \texttt{t1.shipment\_id} instead of the correct column, \texttt{t1.invoice\_number}) has a low SCF of $0.15$. 
        We use multivariate Platt scaling (MPS) to combine SCF scores to produce an overall accurate, calibrated score (not shown in the figure but described in Section~\ref{subsec:mps}). 
    }
    \label{fig:scf_example}
\end{figure}


Text-to-SQL parsing is the problem of translating natural language queries into Structured Query Language (SQL) code. As the scale of structured data continues to grow, languages such as SQL provide a natural way to access the data of interest~\citep{popescu2003towards}. Text-to-SQL parsing therefore aims to provide a natural language interface for users who need not become familiar with the SQL language syntax and data schema~\cite{giordani-moschitti-2012-translating,iyer-etal-2017-learning,zhong2017seq2sql}. 

While large language models (LLMs) have shown improved performance on standard benchmarks~\cite{rai-etal-2023-improving,gao2023text,pourreza2024din}, they can produce erroneous outputs \citep{huang2023survey} while exhibiting high confidence \citep{borji2023categorical, zhou-etal-2024-relying}. Developing real-world systems requires more than just achieving high accuracy on a task---for users to have more control over how an output of the model is used, such systems must possess the ability to provide additional transparency in the system's behavior~\citep{yao-etal-2019-model,wang-etal-2023-know}. Consequently, applying uncertainty quantification to language modeling has become imperative for many NLP systems---including text-to-SQL---in order to communicate the risks associated with treating model outputs as correct~\cite{dong-etal-2018-confidence}.


In this paper, we study uncertainty quantification in the form of calibration, aiming to derive uncertainty (or ``risk'') estimates using \textit{post-hoc} methods, which can be applied to any model output. These estimates take on the form of probabilities, and the goal is for the reported probabilities to match the true probability of the model being correct. 

Deep learning-based models often produce probabilities that are not well-calibrated \citep{guo2017calibration}.
Therefore, we consider a canonical \textit{post-hoc} calibration approach, Platt scaling~\cite{platt1999probabilistic}, and demonstrate that it significantly improves calibration of LLMs on text-to-SQL. Moreover, we then propose an extension of Platt scaling, which we call multivariate Platt scaling (MPS), to further improve performance by exploiting the fact that SQL syntax is highly structured and can be broken down into sub-clauses.
Our intuition is that the frequency of each of the generated sub-clauses across multiple likely output samples from the model (named ``sub-clause frequency'' or SCF) is indicative of confidence (or lack thereof) and can therefore be used in MPS. In Figure~\ref{fig:scf_example}, we present a graphical summary of our approach.

Our contributions are as follows:
\begin{itemize}[noitemsep,topsep=0pt,leftmargin=1em]
    \item We benchmark post-hoc calibration of LLMs on text-to-SQL parsing using a variety of metrics for directly measuring calibration, as well as the related task of error detection---we find that LLMs are very miscalibrated out-of-the-box;
    \item We propose multivariate Platt scaling (MPS), which uses sub-clause frequencies (SCF) from multiple generated samples---our experiments show that MPS consistently outperforms baseline approaches;
    \item We perform analyses of calibration across models, query complexity, and shifts in probability magnitude to identify patterns related to calibration performance.
\end{itemize}


%

\section{Related Work}

\paragraph{Uncertainty Quantification of NLP models. }

Uncertainty quantification has been studied for a few NLP tasks such as machine translation \citep{blatz-etal-2004-confidence,ueffing-ney-2005-word,kumar2019calibration, wang2020inference} and question-answering \citep{gondek2012,10.1162/tacl_a_00407,si2022re, kadavath2022language, lin2023generating}. More recently, \citet{lin2022teaching} and \citet{xiong2023can} explore whether a model can express uncertainty through verbalized confidence.

We note that such works are tangentially related to ours in that they proposed base uncertainty scores. In contrast, our work studies \textit{post-hoc} calibration, which takes base score (like the ones mentioned above) of some LLM output and further calibrates it. As discussed later, given the lack of prior work proposing base scores specifically for text-to-SQL, we calibrate model probabilities directly. However, in Table \ref{tab:base_scores} of the appendix, we test other general base uncertainty scores for LLMs (e.g., perplexity, P(True), verbalized confidence) and find that post-hoc calibration performs roughly the same across them.

Regarding similar works to ours that study post-hoc uncertainty but for other NLP tasks, \citet{detommaso2024multicalibration} study a variation of calibration---\textit{multi}calibration \citep{hebert2018multicalibration}---for question-answering. Framing long-form text generation as a \textit{conformal prediction} \citep{shafer2008tutorial} problem, \citet{mohri2024language} provide high probability guarantees for factuality. \citet{liu2024multi} extend both works, studying multicalibration and \textit{multivalid} conformal prediction \citep{jung2022batch} for long-form text generation.

\paragraph{Calibration in Semantic Parsing. }

A couple works have studied calibration for semantic parsing tasks. 
\citet{dong-etal-2018-confidence} train a gradient-boosted tree model to predict confidence scores by using features derived from knowledge of the training data or by applying perturbations to the model itself.
\citet{stengel2023calibrated} benchmark how well-calibrated token probabilities from the underlying language model are for text-to-SQL parsing, finding that these probabilities are very miscalibrated.
In contrast, our work establishes how calibrated LLMs are on text-to-SQL \textit{after} correcting model probabilities using post-hoc calibration. Moreover, we propose an improved post-hoc approach that utilizes additional generations from the model.




\paragraph{Additional Related work for Text-to-SQL Parsing.} 

We propose a method for calibration that partially relies on extracting additional information from sub-clauses in SQL outputs. Similarly, past works have also used the structure of SQL syntax for various purposes in text-to-SQL parsing. For example, \citet{chen2023text} propose a SQL query correction method that considers edits at the clause level, and \citet{tai2023exploring} adapt chain-of-thought prompting for text-to-SQL, constructing reasoning paths based on the logical execution order of SQL clauses. For error detection, which we use as an auxiliary evaluation metric that supplements our calibration study, \citet{chen2023error} build an error detection model by training on realistic parsing errors that they collect and annotate from model outputs. Finally, \citet{wang-etal-2023-know} propose methods for handling ambiguous and unanswerable questions for text-to-SQL models.
\section{Calibration}

Calibration refers to an alignment between reported and observed probabilities. To define calibration formally, we have to first define the event of interest for which probabilities are being reported. In the context of text-to-SQL parsing using LLMs, we consider the problem of producing calibrated probabilities for the correctness of the query that has been produced by the LLM. Adapting standard notation for statistical calibration to our problem, we have that a sample looks like $(X,Y) \sim \mathcal{D}$, where $X \in \mathcal{X}$ is the SQL query output by the LLM and $Y \in \{0, 1\}$ indicates the correctness of the SQL query corresponding to the natural language prompt (which typically include both the question and database schema) fed into the LLM : \[Y = \mathbf{1}\{X \text{ is accurate w.r.t the prompt}\}.\]

Our goal is to produce an uncertainty score function $s: \mathcal{X} \to [0, 1]$ that measures the confidence of $X$ being correct. While a basic requirement for $s$ is that higher values should denote higher levels of confidence, we can also ask if $s$ is calibrated, meaning that it satisfies,
\begin{equation}\label{eq:perfect}
    P_{\mathcal{D}} (Y=1 \mid s(X) = p) = p, \forall x \in \mathcal{X}.
\end{equation}
In other words, when the reported probability or confidence score is $p$, the probability of the SQL query being correct is also $p$.

However, because $p$ can take on an infinite number of values between $0$ and $1$, ensuring that Equation \ref{eq:perfect} holds for all $p$ is intractable \citep{gupta2020distribution}. A typical simplification is to consider calibration when the output of $s$ is discretized into coarser probability bins (e.g., $[0. 0.2), \ldots, [0.8. 1.0])$ and ensuring calibration on those bins. This is the definition we will aim for. 
\begin{definition}\label{def:calibration}
(Calibration) Suppose we have $k$ probability bins $S_i$ that partition $[0,1]$. Then a scoring function $s$ is calibrated w.r.t $\mathcal{D}$ if
\begin{equation}
    \Delta_i(s) = 0, \forall i \in 1 \ldots k,
\end{equation}
where $\Delta_i(s)$ is the bias of $s$ for the $k$-th bin $S_i$:
\begin{equation}\label{eq:bias}
    \Delta_i(s) = \mathbb{E}_{\mathcal{X \sim D}}[Y - s(X) \mid s(X) \in S_i].
\end{equation}
\end{definition}
%
For further background on binning approaches to calibration, see \citet{gupta2022post}.

We highlight that in this framework, we do not need to conduct additional modeling on the natural language prompt. The prompt goes through the LLM to give us the output SQL query $X$ that is validated ($Y$) according to the prompt. Then afterwards, it is no longer used for calibration.

\section{Methods}

Our methodological contributions are (1) the method of multivariate Platt scaling and (2) its adaptation to semantic parsing using extracted sub-clause frequencies. We highlight that our approach to uncertainty quantification for text-to-SQL, which is motivated by high-level statistical insights, are novel and likely have implications for other tasks in NLP and machine learning in general.

\subsection{Platt Scaling}

Platt scaling \citep{platt1999probabilistic} is a technique that calibrates the probabilistic outputs of a model using a logistic regression mapping. Given a score $s \in [0,1]$ produced using some model, canonical Platt scaling corresponds to learning two weights, $w_0, w_1 \in \mathbb{R}$ that map $s$ to a new value in $[0,1]$:
\begin{equation}
    s_\text{PS} \mapsto \text{sigmoid}(w_0 + w_1\cdot \text{logit}(s)).\label{eq:platt-basic}
\end{equation}
Here $\text{sigmoid}(x) = \tfrac{1}{1+e^{-x}}$ and $\text{logit}(x) = \log\left(\tfrac{x}{1-x}\right)$. These are inverses, so setting $(w_0, w_1) = (0, 1)$ recovers the identity mapping.\footnote{Other values of $(w_0, w_1)$ correspond to natural monotonic mappings that can be used for calibration; see \citet[Figure~1 and others]{niculescu2005predicting} for examples.} These weights are typically learned on held out data such as the validation data. We discuss this further in the following subsection. 

Platt scaling is simple, interpretable, and often works well in practice. Many variants of Platt scaling have recently been developed: for calibrating the top-label probabilities of deep nets \citep{guo2017calibration, gupta2021top}, calibrating full-vector probabilities in multiclass settings \citep{kull2019beyond}, and combining calibration with online learning style adversarial (worst-case) guarantees \citep{gupta2023online}. 

In addition, we note that the related method of temperature scaling \citep{guo2017calibration, kull2019beyond} was introduced primarily as a way of conducting post-hoc calibration for \textit{multi-class} classification. We highlight, however, that temperature scaling and Platt scaling are equivalent for binary classification (i.e., our setting of determining whether a SQL output is correct or not). Specifically, temperature scaling can be written down as a special case of Platt scaling such that weights $w_0 = 0$ and $w_1 = \frac{1}{\tau}$, where $\tau$ is the temperature learned in temperature scaling.

\subsection{Multivariate Platt Scaling}\label{sec:platt}
\label{subsec:mps}
Our work provides a multivariate extension of Platt scaling (MPS) that is natural, yet understudied (to the best of our knowledge). Suppose a number of signals $s_1, s_2, \ldots, s_m \in \mathbb{R}$ are available for predicting a single probability in $[0,1]$. Given these signals, we learn (in the spirit of Equation~\ref{eq:platt-basic}) $m + 1$ weights $w_0, w_1, \ldots, w_m \in \mathbb{R}$ that correspond to the mapping,
\begin{equation}
    s_\text{MPS} \mapsto \text{sigmoid}(w_0 + w_1s_1 + \ldots + w_ms_m),\label{eq:platt-multivariate}
\end{equation}
so that $s_\text{MPS} \in [0,1]$ is the final, combined signal.


Although Equation \ref{eq:platt-multivariate} resembles typical ML models, we emphasize a key difference in that while signals $s_i$ are learned on the training data, \textbf{the weights $\mathbf{w_i}$ are learned on a held out validation set}---also called \textbf{calibration} data. Typically, the validation set is only used for model selection and not modeling itself so that we have some out-of-sample metric that checks for overfitting. In our case, since the secondary Platt scaling model is small ($m+1$ parameters), we can reuse validation data \textit{without significant overfitting}, as long as a reasonable number of validation points are available.\footnote{say $\approx 10m$, which is based on the sample complexity of logistic regression (see \citet{hsu2024sample}).} Multivariate Platt scaling is one of many ``post-hoc" calibration approaches that aim to improve calibration of a model without sacrificing accuracy. 


Upon reviewing the literature, we found only a couple of instances where an approach similar to MPS is used. 
In work on object detection in images, \citet{kuppers2020multivariate} utilize predicted bounding box locations as additional signals for calibration.
In addition, \citet{gopalan2022low} adapt Platt scaling to \textit{multi}-calibration, where the goal is to be calibrated while conditioning on predefined subgroups. In their formulation, $s_i$ are instead binary features that denote whether some data point belongs to a group.
In contrast, (1) high-level statistical insight discussed in the preceding paragraph and (2) its application to uncertainty quantification for text-to-SQL are novel, and likely have implications for other problems, especially in the context of semantic parsing where additional signals of correctness may be present in the model.

\subsection{Extracting Sub-Clause Frequencies}\label{sec:methods_scf}

We now describe how the signals $s_i$ are computed for the task of text-to-SQL parsing. Generally, one can obtain many signals from LLMs aside from the probability of the full sequence output. In our work, we leverage the fact that the output is structured, comprising multiple sub-clauses for which we can compute output frequencies.

Our method draws inspiration from \textit{self-consistency} \citep{wang2022self} for a model that produces multiple outputs for the same prompt. Specifically, each candidate output is assigned a frequency score that counts the number of times the candidate appears among all outputs. The one with the highest frequency score (i.e., the most consistently generated candidate) is selected as the final output.

In our method, we too calculate frequency scores, but instead of using the scores to choose the best response, we use them as measurements of uncertainty. Moreover, we exploit the fact that SQL queries are structured in a particular way. For example, all queries must contain \texttt{SELECT} and \texttt{FROM} clauses, and there exist only a finite number of additional clause types that a query may include (e.g., \texttt{WHERE}, \texttt{GROUP BY}, etc.). In addition, while a theoretically infinite number of SQL queries can be composed together (via operators such as \texttt{JOIN}, \texttt{UNION}, etc.), the number of sub-queries found in most SQL queries (including those in prevalent text-to-SQL benchmarks) is small. Thus, instead of calculating the number of times the entire SQL query output appears, as is typically done in \textit{self-consistency}-based methods, our method counts the number of times each clause appears.

\begin{figure}[tbp!]
    \centering
    \includegraphics[width=0.8\columnwidth]{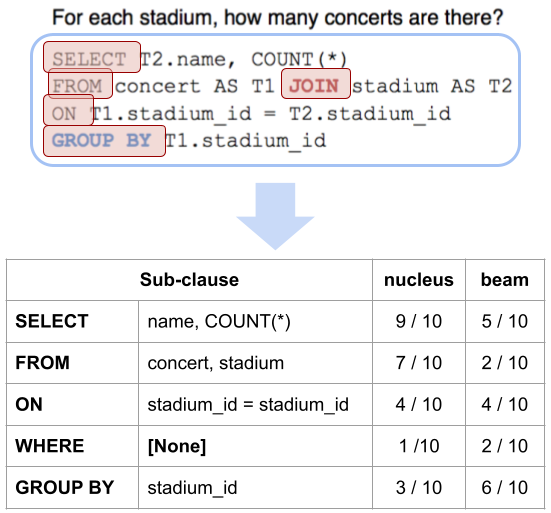}
    \caption{
        We show how to parse a query from the SPIDER dataset in order to derive sub-clause frequency scores $s_i$. Specifically, we count the number of times each sub-clause appears among outputs derived from (1) nucleus sampling and (2) beam search. Note that for illustrative purposes, we show a simplified example in which we assume in our data universe that we do not have multiple \texttt{SELECT} statements composed together and the only possible sub-clauses that can appear are \texttt{SELECT}, \texttt{FROM}, \texttt{ON}, \texttt{WHERE}, and \texttt{GROUP BY}. In this example, we have in total, $10$ signals $s_i$ that can be provided to our method, MPS.
    }
    \label{fig:parsing_example}
\end{figure}

As a result, we produce signals $s_i$ for every sub-clause, to which we apply multivariate Platt scaling as described in the previous subsection. In Figure \ref{fig:parsing_example}, we provide a simplified example of what this procedure looks like using nucleus sampling \citep{holtzmancurious} and beam search \citep{freitag2017beam}. Appendix \ref{appx:additional_scf} describes our procedure in full detail.

\section{Experimental Setup}

We note that for all experiments, we use the calibration set (defined in Section \ref{sec:datasets}) to construct our mappings $s_{\text{PS}}$ and $s_{\text{MPS}}$. We then evaluate our mappings on the test sets for each dataset, reporting our findings over a single run of the calibration methods (without hyperparameter tuning).

\subsection{Datasets}\label{sec:datasets}

We evaluate our methods on two popular benchmarks for (English) text-to-SQL parsing: SPIDER \citep{yu2018spider} and BIRD \citep{li2024can}.

SPIDER consists of 10,181 questions and 5,693 unique complex SQL queries on 200 databases covering 138 domains. Serving as a complex, cross-domain dataset, SPIDER helped define a new semantic parsing benchmark in which databases at test time are not seen during training. As a result, models are tasked to generalize to new database schemas, which are provided alongside natural language questions during evaluation. The \textbf{test} set has 2,148 examples, and the development set (which we use as our \textbf{calibration} set) has 1,034 examples.

BIRD consists of 12,751 unique question-SQL pairs on 95 databases, covering 37 domains. Focusing on the challenge of handling dirty and noisy database values, the queries in BIRD are often more complex than those in SPIDER. In addition, \citet{li2024can} introduce the notion of external knowledge, which contains information mapping natural language questions to database values. Each question is paired with some external knowledge to provide additional assistance to LLMs parsing the question. Because a test set has not yet been released publicly, we use the development set (of size 1,534) as our \textbf{test} set and randomly sample 1,000 examples from the train set to use as our \textbf{calibration} set.

\subsection{Models}

We use \textsc{T5-3B-PICARD} (which we will refer to as \textbf{\textsc{T5 3B}} going forward) and \textbf{\textsc{Llama 3.1 8B Instruct}} on the SPIDER dataset. The \textsc{T5-3B-PICARD} model is a T5 model (available at this \href{https://huggingface.co/tscholak/cxmefzzi}{link}) fine-tuned on SPIDER using constrained decoding for text-to-SQL parsing, achieving strong results on the task \citep{scholak2021picard}. \textsc{Llama 3.1 8B Instruct}~\citep{dubey2024llama} is used with zero-shot prompting because the model is also able to obtain strong accuracy on this task.
For the BIRD dataset, we use both \textbf{\textsc{Llama 3.1 8B Instruct}} and \textbf{\textsc{Llama 3.1 70B Instruct}}~\citep{dubey2024llama} with zero-shot prompting as they also have been shown to perform relatively well. We did not use T5 variants for this dataset as their accuracies were low ($23.34\%$) \citep{li2024can}.
Prompts for our experiments are presented in Appendix \ref{appx:additional_exp_details}.

We note that our experiments use open-weight models because we have access to token probabilities and are able to control the resampling of outputs, which is not possible using closed-source models like GPT-4o. In addition, open-weight models facilitate reproducibility of our findings.

In addition, as mentioned in Section~\ref{sec:methods_scf}, our calibration approaches rely on generating multiple outputs for each question in our datasets. For each set of generations, we take the output with the highest model probability (sum over log token probabilities given by the model) as our primary prediction. The probability for the primary prediction is then relearned (calibrated) using the techniques of this paper. Table~\ref{tab:model_acc} in the Appendix shows the accuracy of the primary prediction on two datasets. 

\subsection{Computing Sub-Clause Frequencies}

To run MPS using parsed sub-clause frequencies as signals, we sample outputs via nucleus sampling and beam search. 
In our experiments, we set the number of samples for each method to $k=10$,\footnote{These numbers were set after initial experiments on the development set. We note that we did not attempt to further tune these hyperparameters, which can only lead to better results for our proposed approach.}
leading to a total of 20 samples. Beam search takes the top $k=10$ outputs by probability (after length normalization) using a beam width of $2k$. Then, we compute sub-clause frequencies for each sample using all samples from nucleus sampling and beam search (as illustrated in Figure \ref{fig:parsing_example}).

We note that we remove a small proportion of outputs that have syntactic errors since the correctness of such outputs can be determined quite easily with automated SQL parsers. 

\subsection{Metrics for Calibration}\label{sec:metrics}

We now present the evaluation metrics we use to measure calibration. Specifically, we consider expected calibration error, adaptive calibration error, and Brier score together in order to provide a more holistic and robust evaluation of calibration.

First, given some set of probability bins $S_i$, we define $\ell_1$-calibration error as the following:
\begin{equation}\label{eq:calibration_error}
    \ell_1\text{-CE}(s) = \sum_{i=1}^k P_{\mathcal{X \sim D}}(s(X) \in S_i) \left| \Delta_i(s) \right| .
\end{equation}
(Recall the definition of $\Delta_i$ from Definition~\ref{def:calibration}.) As Equation \ref{eq:calibration_error} suggests, calibration error is highly dependent on the choice of probability bins $S_i$. We consider two variations:
\begin{enumerate}
    \item \textbf{\textit{expected} calibration error (ECE)} \citep{naeini2015obtaining}: bins $S_i$ are defined as $k$ equal-width bins: $[0, \frac{1}{k}), [\frac{1}{k}, \frac{2}{k}), \ldots, [\frac{k-1}{k}, 1]$.
    
    \item \textbf{\textit{adaptive} calibration error (ACE)}  \citep{Nixon_2019_CVPR_Workshops}: bins $S_i$ are defined such that the all sizes of all bins are the same (i.e., $P_{\mathcal{X \sim D}}(s(X) \in S_i) = \frac{1}{k}$).
\end{enumerate}
%

In addition to minimizing calibration error, it is often desirable to output a wide range of scores $s$ that can help users better distinguish between high and low confidence outputs. The spread of scores is sometimes referred to as \textit{resolution} \citep{brocker2009reliability} in the literature, where higher resolution corresponds to a larger spread. Calibration error, however, does not capture this spread. For example, as an extreme case, an algorithm could output the same probability $s(x) = P_D(Y=1)$ for all examples in $x \in D$. Despite being perfectly calibrated (i.e., calibration error of $0$), this scoring function serves no purpose in communicating risk for individual examples $X$.

Consequently, another metric that is often considered for evaluating calibration is \textbf{Brier score}, which is the mean squared error of $s(x)$
\begin{equation}\label{eq:brier_score}
    BS(s) = \mathbb{E}_{\mathcal{X \sim D}}[ (Y - s(X))^2]
\end{equation}
Intuitively, a lower Brier score is desirable since correct outputs ($Y=1$) should be accompanied by higher probability estimates ($s(X)$ closer to $1$) and incorrect outputs ($Y=0$) by lower probability estimates ($s(X)$ closer to $0$). Moreover, unlike calibration error, Brier score penalizes algorithms with low resolution.\footnote{In fact, as shown in \citet[Equation~13]{brocker2009reliability}, Brier score can be decomposed into calibration error and (negative) resolution, thereby capturing both properties that are valuable for uncertainty quantification.}
\section{Results}\label{sec:results}

\subsection{Calibration}

\begin{table*}[t!]
\centering
\begin{tabular}{c c l c c c c}
\toprule
\textbf{Dataset} & \textbf{Model} & \textbf{Method} 
& \textbf{Brier} ($\downarrow$) & \textbf{ECE} ($\downarrow$) & \textbf{ACE} ($\downarrow$) & \textbf{AUC} ($\uparrow$) \\ 
\toprule

\multirow{6}{*}{SPIDER}

& \multirow{3}{*}{\textsc{T5 3B}}                    
&   Uncalibrated         & 0.1893           & 0.1632           & 0.1632          & 0.7200               \\ 
& & PS                   & 0.1656           & 0.0623           & 0.0628          & 0.7200          \\ 
& & MPS (\textit{ours})  & \textbf{0.1566}  & \textbf{0.0264}  & \textbf{0.0253} & \textbf{0.7785} \\

\cmidrule{2-7}

& \multirow{3}{*}{\textsc{Llama 3.1 8B Instruct}}
&   Uncalibrated         & 0.3576           & 0.4161           & 0.4147          & 0.7123                \\ 
& & PS                   & 0.1726           & 0.0437           & 0.0449          & 0.7123          \\ 
& & MPS (\textit{ours})  & \textbf{0.1626}  & \textbf{0.0244}  & \textbf{0.0260} & \textbf{0.7475} \\ 

\midrule

\multirow{6}{*}{BIRD}

& \multirow{3}{*}{\textsc{Llama 3.1 8B Instruct}}
&   Uncalibrated         & 0.2378           & 0.1722           & 0.1688          & 0.6913          \\ 
& & PS                   & 0.2129           & 0.0556           & 0.0739          & 0.6913          \\ 
& & MPS (\textit{ours})  & \textbf{0.1950}  & \textbf{0.0418}  & \textbf{0.0423} & \textbf{0.7353} \\ 

\cmidrule{2-7}

& \multirow{3}{*}{\textsc{Llama 3.1 70B Instruct}}
&   Uncalibrated         & 0.2866           & 0.2343           & 0.2324          & 0.6650              \\ 
& & PS                   & 0.2316           & 0.0549           & 0.0631          & 0.6650          \\ 
& & MPS (\textit{ours})  & \textbf{0.2141}  & \textbf{0.0314}  & \textbf{0.0323} & \textbf{0.7173} \\ 

\bottomrule
\end{tabular}
\caption{
We report Brier score, expected calibration error (ECE), and adaptive calibration Error (ACE) on the SPIDER and BIRD datasets. In addition, we report AUC for the task of error detection. These metrics are calculated after applying Platt scaling (\textbf{PS}, \textit{baseline}) to token probabilities and Multiplicative Platt scaling (\textbf{MPS}, \textit{ours}) to token probabilities and parsed frequencies (derived from nucleus sampling and beam search). For reference, we also report (\textbf{Uncalibrated}) how well calibrated the model token probabilities are prior to applying Platt scaling.
}
\label{tab:calibration+auc}
\end{table*}

Table \ref{tab:calibration+auc} presents the results of our LLM-based text-to-SQL parsers on both datasets using the two approaches to calibration: Platt scaling (PS) and multivariate Platt scaling (MPS) with sub-clause frequencies (SCF). 
Here, MPS uses SCF scores derived from both nucleus sampling and beam search.
%
As reference, we report results for uncalibrated probabilities \citep{stengel2023calibrated} that are generated directly from the model.

We observe that the probabilities produced by the LLMs are consistently and significantly miscalibrated across all experiments. However, Platt scaling (PS) improves calibration across both datasets and models. In some cases, PS improves calibration metrics by very wide margins (e.g., \textsc{Llama 3.1 8B Instruct} on SPIDER shows a reduction of $0.17$ in Brier score and $0.37$ in ECE and ACE).

Our method (MPS) further improves performance of calibration, cutting calibration error by over $50\%$ in some cases (e.g., \textsc{T5 3B} on SPIDER). We highlight that MPS maintains this advantage across all datasets and models, providing strong evidence for the utility of using the sub-clause structure of SQL syntax to derive additional signals $s_i$ for producing calibrated probabilities. 



\subsection{Evaluating Error Detection}

We evaluate our methods for calibration on the related task of error detection to demonstrate their additional utility. In particular, we assess how the (calibrated) probabilities derived from MPS can be used to detect errors in the model outputs. Using AUC as our evaluation metric, our results characterize the trade-offs between true and false positive rates when choosing different thresholds for determining if a SQL output is erroneous (e.g., one could predict that a SQL output is wrong whenever MPS produces a probability $\le 0.3$). In Table \ref{tab:calibration+auc}, we again show that MPS using SCF performs the best, achieving higher AUC across all datasets and models. We note that because PS applies a monotonic transformation to the uncalibrated model probabilities, it---unlike MPS---can only affect calibration but not error detection (i.e., AUC remains unchanged after applying PS).

\section{Analysis}\label{sec:analysis}

\subsection{Calibration Plots}

\begin{figure*}[t!]
    \centering
    \includegraphics[width=1.0\textwidth]{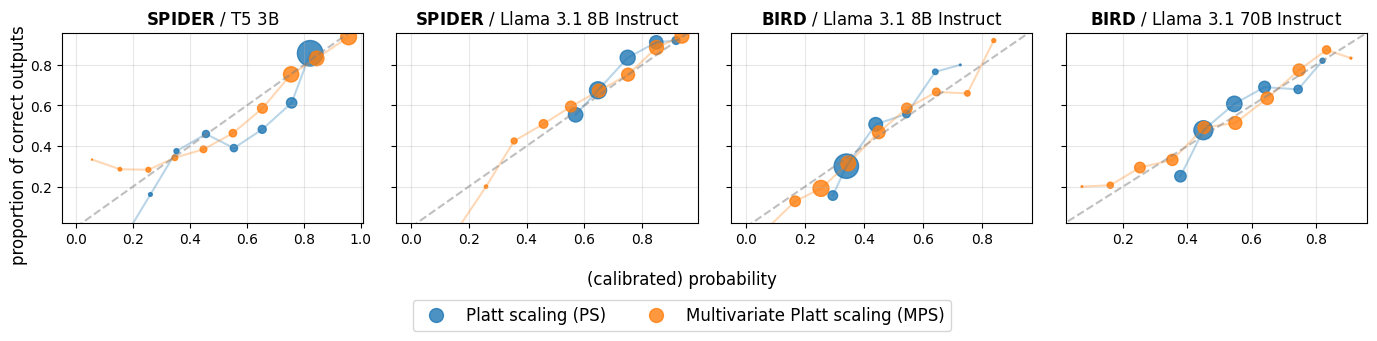}
    \caption{
        Calibration curves (using fixed-width ECE-style bins) comparing Platt scaling ({\textcolor{blue}{blue}}) applied to final token probabilities, and (ours) multivariate Platt scaling ({\textcolor{orange}{orange}}) applied to token probabilities and sub-clause frequencies. The radius of the markers in the scatter plot are proportional to the number of points in the corresponding probability bin. For instance, the blue marker near $(0.8, 0.8)$ of the top-left plot corresponds to $67.4\%$ of the data with predicted probabilities in the range $[0.8, 0.9)$. Across all models (T5, Llama) and datasets (SPIDER, BIRD), our method is better calibrated (scatter plot close to the $x=y$ line), and has higher \textit{resolution} (varied values on the $x$-axis).
    }
    \label{fig:ece}
\end{figure*}

Figure \ref{fig:ece} shows the calibration plots using ECE, comparing probabilities calibrated by PS and MPS. To convey the number of examples in each bin, we scale the radius of each marker by the bin size. Generally, we observe that calibration errors are closer to $0$ (i.e., points lie closer to $y=x$) for MPS compared to PS, reflecting the lower ECE of MPS.

In addition, we examine qualitatively the spread of the probabilities produced by PS and MPS. In most cases, a more spread out or higher \textit{resolution} forecast is more desirable for uncertainty quantification and is correlated with smaller Brier scores (see Section \ref{sec:metrics}). For the SPIDER dataset, we observe that probabilities produced by PS with \textsc{T5 3B} are heavily concentrated around $0.8$. PS with \textsc{Llama 3.1 8B Instruct} only outputs probabilities $\ge 0.5$. Similarly, on the BIRD dataset, PS with \textsc{Llama 3.1 8B Instruct} mostly outputs probabilities around a single value ($0.3$), while for PS with \textsc{Llama 3.1 70B Instruct}, most probabilities produced by PS fall between $0.4$ and $0.7$. In contrast, we observe that MPS consistently produces probabilities that are more spread out (while still lying close to $y=x$).

\begin{figure*}[htbp!]
    \centering
    \includegraphics[width=1.0\textwidth]{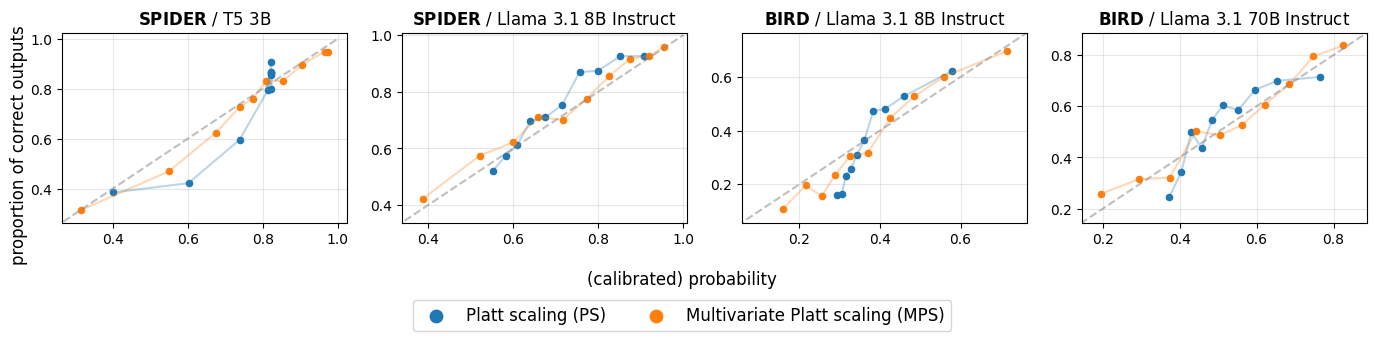}
    \caption{
        Calibration curves (using \textbf{ACE}) comparing (baseline) Platt scaling (PS, {\textcolor{blue}{blue}}) applied to final token probabilities, and (ours) multivariate Platt scaling (MPS, {\textcolor{orange}{orange}}) applied to token probabilities and sub-clause frequencies. Across all models (T5, Llama) and datasets (SPIDER, BIRD), our method is better calibrated (scatter plot close to the $x=y$ line), and has higher \textit{resolution} (varied values on the $x$-axis).
    }
    \label{fig:ace}
\end{figure*}

In Figure \ref{fig:ace}, we plot adaptive calibration error, where each point (or bin) represents roughly the same number of samples. Qualitatively, these plots are similar to those in Figure \ref{fig:ece} and thus lead to similar findings---our method, MPS, is better calibrated (scatter plot close to the $x=y$ line) and has higher resolution (varied values on the $x$-axis) when compared to PS.



\subsection{Comparing Calibration across Models}\label{sec:analysis_compare_models}

When examining (uncalibrated) probabilities generated directly by the LLM on SPIDER, we observe that those produced by fine-tuned \textsc{T5 3B} are better calibrated (i.e., Brier score, ECE, and ACE are almost $2$x smaller) than those of the larger \textsc{Llama 3.1 8B Instruct} model, despite the two models achieving similar test accuracies (Table \ref{tab:model_acc}). However, after applying (multivariate) Platt scaling, the differences become small, providing strong evidence that when evaluating calibration for LLMs, post-hoc calibration (e.g., Platt scaling) should be conducted before forming conclusions about how well-calibrated LLMs are relative to each other.


On BIRD, Table \ref{tab:calibration+auc} reports that the smaller \textsc{Llama 3.1 8B Instruct model} is better calibrated w.r.t. Brier score than its larger \textsc{70B} counterpart, both before and after applying (multivariate) Platt scaling. We stress, however, that Brier score cannot be compared across models that achieve such different test accuracies (Table \ref{tab:model_acc}) since Brier score is very sensitive to this value. For example, the base rates (i.e., the Brier score when simply outputting the proportion of correct outputs $P(Y=1)$ for all data points) for \textsc{Llama 3.1 8B Instruct} and \textsc{Llama 3.1 70B Instruct} are $0.2300$ and $0.2489$ respectively, and so one might expect that the Brier score of the 8B variant to be lower, post-calibration. 




\subsection{Ablation Study for Deriving SCF Scores}\label{sec:ablation_scf}

We run ablation studies to identify the impact of using different methods for computing the outputs used to calculate the SCF scores: nucleus sampling only, beam search only, or both (which corresponds to the results for MPS presented in Table~\ref{tab:calibration+auc}). We emphasize that regardless of this choice, MPS outperforms PS in all cases.

The results, available in full in Table \ref{tab:calibration+auc_scf} in the Appendix, show that no single sampling method is better than the other, with different methods being best depending on the dataset and LLM used. We find that using both provides the most well-rounded performance across all experiments. However, we stress that data and model specific optimal performances could be obtained if one were to treat signal $s_i$ selection as a hyperparameter that can be tuned. 

\subsection{Sensitivity analysis for MPS}

In Table \ref{tab:vary_calibration_set}, we show the performance of MPS with the size of the calibration set. We see that the performance of MPS declines as the size of the calibration set decreases.

Next, we study the impact of the number of model outputs $k$ used to calculate SCF scores for MPS in~\ref{tab:vary_k}. As expected, the larger number of outputs used increases performance of calibration, with a noticeable increase for more than 1 output and with $k=5$ as used in the experiments in this paper being a good trade-off between latency and performance and larger values bringing additional improvements.

\begin{table}[ht!]
\centering
\resizebox{\columnwidth}{!}{
\begin{tabular}{cc|cccc}
\toprule
\textbf{dataset} & \textbf{factor} & \textbf{Brier} & \textbf{ECE} & \textbf{ACE} & \textbf{AUC} \\
\toprule
\multirow{7}{*}{SPIDER}
 & all   & 0.164 & 0.023 & 0.023 & 0.746 \\
 & $2^{-1}$  & 0.166 & 0.033 & 0.036 & 0.741 \\
 & $2^{-2}$  & 0.168 & 0.036 & 0.039 & 0.737 \\
 & $2^{-3}$  & 0.173 & 0.047 & 0.050 & 0.725 \\
 & $2^{-4}$  & 0.175 & 0.045 & 0.048 & 0.713 \\
 & $2^{-5}$  & 0.180 & 0.060 & 0.063 & 0.691 \\
 & $2^{-6}$  & 0.192 & 0.094 & 0.099 & 0.637 \\
\midrule
\multirow{7}{*}{BIRD}
 & all   & 0.194 & 0.035 & 0.032 & 0.737 \\
 & $2^{-1}$  & 0.197 & 0.035 & 0.036 & 0.727 \\
 & $2^{-2}$  & 0.200 & 0.042 & 0.043 & 0.717 \\
 & $2^{-3}$  & 0.206 & 0.056 & 0.055 & 0.701 \\
 & $2^{-4}$  & 0.212 & 0.065 & 0.066 & 0.686 \\
 & $2^{-5}$  & 0.227 & 0.106 & 0.106 & 0.668 \\
 & $2^{-6}$  & 0.243 & 0.141 & 0.141 & 0.627 \\
\bottomrule
\end{tabular}
}
\caption{
We evaluate MPS on outputs from \textsc{Llama 3.1 8B Instruct} on SPIDER and BIRD when reducing the size of the calibration set. We sample a \textbf{factor} of the original calibration set to run MPS and evaluate on the test set, taking an average over 10 random runs.
}
\label{tab:vary_calibration_set}
\end{table}


\begin{table}[ht!]
\centering
\resizebox{\columnwidth}{!}{
\begin{tabular}{cc|cccc}
\toprule
\textbf{Dataset} & \textbf{k} & \textbf{Brier} & \textbf{ECE} & \textbf{ACE} & \textbf{AUC} \\
\toprule
\multirow{4}{*}{SPIDER} 
 & 10 & 0.164 & 0.023 & 0.023 & 0.746 \\
 & 5  & 0.167 & 0.027 & 0.030 & 0.732 \\
 & 3  & 0.168 & 0.028 & 0.025 & 0.726 \\
 & 1  & 0.170 & 0.040 & 0.038 & 0.717 \\
\midrule
\multirow{4}{*}{BIRD} 
 & 10 & 0.194 & 0.035 & 0.032 & 0.737 \\
 & 5  & 0.194 & 0.041 & 0.035 & 0.732 \\
 & 3  & 0.195 & 0.046 & 0.041 & 0.728 \\
 & 1  & 0.203 & 0.029 & 0.041 & 0.695 \\
\bottomrule
\end{tabular}
}
\caption{
We evaluate MPS on outputs from \textsc{Llama 3.1 8B Instruct} on SPIDER and BIRD when reducing the number of additional model outputs $k$ that are used to calculate SCF scores.
}
\label{tab:vary_k}
\end{table}

\subsection{Uncertainty scores used for post-hoc calibration}

We study additional methods for computing the base uncertainty scored used for calibration and consider specifically perplexity, P(True) \citep{kadavath2022language}, and verbalized confidence \citep{tian2023just}. We ran Platt scaling (PS) on additional base uncertainty scores using outputs from \textsc{Llama 3.1 8B Instruct} on the SPIDER dataset and present the results in Table~\ref{tab:base_scores}.

While applying PS to verbalized confidence performs poorly, PS is able to achieve low calibration error on the other base scores. However Brier score and AUC vary greatly. For example, PS on perplexity performs poorly w.r.t to Brier score and AUC but the best w.r.t ECE and ACE. Finally, we stress that MPS using SCF scores (Table \ref{tab:calibration+auc}) still performs better than all these configurations of PS, highlighting the effectiveness of our proposed method.

\begin{table}[h!]
\centering
\resizebox{\columnwidth}{!}{
\begin{tabular}{l | c c c c}
\toprule
\textbf{Base score} & \textbf{Brier} & \textbf{ECE} & \textbf{ACE} & \textbf{AUC} \\
\toprule
Model Probabilities & \textbf{0.173} & 0.044 & 0.045 & \textbf{0.712} \\
Perplexity & 0.185 & \textbf{0.022} & \textbf{0.028} & 0.613 \\
P(True) & 0.177 & 0.027 & 0.037 & 0.678 \\
Verbalized Confidence & 0.19 & 0.034 & 0.052 & 0.564 \\
\bottomrule
\end{tabular}
}
\caption{
We run Platt Scaling on various base scoring functions on outputs from \textsc{Llama 3.1 8B Instruct} on the SPIDER dataset. In general, we find that aside from when using verbalized confidence, Platt Scaling performs similarly across base scores. We stress, however, that using combining MPS with model probabilities outperforms PS in across all metrics. 
}
\label{tab:base_scores}
\end{table}



\section{Conclusion}

Uncertainty quantification remains an open, yet crucial problem for LLMs. In this paper, we study calibration for text-to-SQL parsing, with a focus on measuring the effectiveness of post-hoc calibration methods (in contrast to previous works like \citet{stengel2023calibrated}). We introduce a method, multivariate Platt scaling,
that leverages the structured nature of outputs for this task and uses sub-clause frequency scores as inputs. Our experiments demonstrate that this method significantly gives better calibrated probabilities for the correctness of the query output by the LLM than the baseline approaches.
Furthermore, we performed several analyses to show that calibration has a large, positive impact on the output of LLMs for this semantic parsing task. 
We hope that this work will inspire future research and developments in this domain, both in terms incorporating structured outputs in uncertainty quantification and for the purpose of improving calibration of text-to-SQL modeling as a whole.


    

\clearpage

\section{Limitations} 

First, the datasets used in our experiments are solely in English, although our methods are applicable to other languages directly as long as a large language model covering the target language is available. This choice allows for consistency and comparability across the datasets, but it does not test the generalizability of our findings to other languages. In future work, we plan to extend our research to a multilingual setting to address this limitation, for example using the multi-lingual SPIDER dataset~\cite{jose2023multilingual}.

The proposed method for calibration requires generating multiple candidate output SQL queries given a natural language input. This generation introduces an increase in inference time, although sampling can be performed in parallel, limiting the potential latency of the response if calibration was included in a real-world application.

The experiments with the \textsc{Llama 3.1} models are limited to zero-shot prompts. We did not experiment with additional fine-tuning or instruction-tuning with text-to-SQL task data, as the models already obtain good performance on the test set and were potentially already trained with data for this task. However, our approach can easily be adopted and used with such models, as shown in experiments with the T5-3B model trained using SPIDER.

\section{Ethical Considerations}

We use publicly available datasets intended for the task of text-to-SQL semantic parsing. The datasets and large language models we used have permissive licenses allowing for research use. We do not envision any potential risks associated with the task discussed in this paper.

\section{Acknowledgements}

This research is supported by a grant from the Bloomberg Data Science Ph.D. Fellowship Program to the first author.

\bibliography{main}

\newpage

\appendix

\onecolumn

\section{Supplementary experiments}\label{appx:additional_results}


\paragraph{Accuracy of language models.}

In Table~\ref{tab:model_acc} we show model performance on SPIDER and BIRD. For our experiments, we sample $10$ outputs using nucleus sampling and $10$ using beam search. Then for each query, we take the model output (among all outputs sampled) with the highest model probability (sum over log token probabilities) as our prediction. Table~\ref{tab:model_acc} shows the performance when choosing outputs in this way, picking the top output among the set of samples generated by (1) nucleus sampling (\textbf{Nucleus}), (2) beam search (\textbf{Beam}), and (3) both (\textbf{N+B}). Generally speaking, we do not see significant different in model accuracy among these three choices and thus choose (3) for our experiments out of convenience. We also note that that \textsc{Llama 3.1 70B Instruct} is a very strong model, performing on par or better on BIRD than various published methods that combine GPT-4 with more advanced prompting strategies \citep{pourreza2024din, gao2023text}.

\begin{table}[h!]
\centering
\begin{tabular}{c c c c c}
\toprule
\textbf{Dataset} & \textbf{Model} & \textbf{Nucleus} & \textbf{Beam} & \textbf{N + B} \\
\toprule

\multirow{2}{*}{SPIDER}
& \textsc{T5 3B}                & 73.52\% & 73.52\% & 73.52\% \\   
& \textsc{Llama 3.1 8B Instr.}  & 74.77\% & 73.77\% & 74.60\% \\       

\midrule

\multirow{2}{*}{BIRD}
& \textsc{Llama 3.1 8B Instr.}  & 34.60\% & 35.65\% & 35.85\% \\   
& \textsc{Llama 3.1 70B Instr.} & 52.28\% & 53.18\% & 53.35\% \\ 

\bottomrule
\end{tabular}
\caption{
We report the accuracy of each model on the corresponding dataset's test set.
}
\label{tab:model_acc}
\end{table}








\paragraph{Varying sampling method for deriving SCF scores.}

As discussed in Section \ref{sec:ablation_scf}, in Table \ref{tab:calibration+auc_scf}, we show how the performance of MPS varies based on the choice of deriving SCF scores (from nucleus sampling only, beam search only, or both).

\begin{table*}[bth!]
\centering
\begin{tabular}{c c c l c c c c}
\toprule
\textbf{Dataset} & \textbf{Model} & & \textbf{Method} & \textbf{Brier} & \textbf{ECE} & \textbf{ACE} & \textbf{AUC} \\ 
\toprule
\multirow{10}{*}{SPIDER}

& \multirow{5}{*}{\textsc{T5 3B}}
&                          & Uncalibrated            & 0.1893           & 0.1632           & 0.1632          & 0.7200           \\ 
\cmidrule{3-8}
& & (PS)                   & Token Probs.            & 0.1656           & 0.0623           & 0.0628          & 0.7200           \\
\cmidrule{3-8}
& & \multirow{3}{*}{(MPS)} & \quad + Nucleus         & 0.1639           & 0.0460           & 0.0538          & 0.7229           \\ 
& &                        & \quad + Beam            & \textbf{0.1565}  & \textbf{0.0241}  & 0.0308          & \textbf{0.7804}  \\ 
& &                        & \quad + N + B           & 0.1566           & 0.0264           & \textbf{0.0253} & 0.7785           \\

\cmidrule{2-8}

& \multirow{5}{*}{\textsc{Llama 3.1 8B Instruct}}
&                          & Uncalibrated            & 0.3576           & 0.4161           & 0.4147          & 0.7123           \\ 
\cmidrule{3-8}
& & (PS)                   & Token Probs.            & 0.1726           & 0.0437           & 0.0449          & 0.7123           \\
\cmidrule{3-8}
& & \multirow{3}{*}{(MPS)} & \quad + Nucleus         & \textbf{0.1625}  & 0.0280           & 0.0360          & \textbf{0.7506}  \\ 
& &                        & \quad + Beam            & 0.1687           & \textbf{0.0244}  & \textbf{0.0230} & 0.7194           \\ 
& &                        & \quad + N + B           & 0.1626           & \textbf{0.0244}  & 0.0260          & 0.7475           \\

\midrule

\multirow{10}{*}{BIRD}

& \multirow{5}{*}{\textsc{Llama 3.1 8B Instruct}}
&                          & Uncalibrated            & 0.2378           & 0.1722           & 0.1688           & 0.6913           \\ 
\cmidrule{3-8}
& & (PS)                   & Token Probs.            & 0.2129           & 0.0556           & 0.0739           & 0.6913          \\ 
\cmidrule{3-8}
& & \multirow{3}{*}{(MPS)} & \quad + Nucleus         & 0.1976           & 0.0574           & 0.0542           & 0.7307          \\ 
& &                        & \quad + Beam            & 0.1989           & 0.0425           & 0.0442           & 0.7213          \\ 
& &                        & \quad + N + B           & \textbf{0.1951}  & \textbf{0.0418}  & \textbf{0.0423}  & \textbf{0.7353} \\

\cmidrule{2-8}

& \multirow{5}{*}{\textsc{Llama 3.1 70B Instruct}}
&                          & Uncalibrated            & 0.2866           & 0.2343           & 0.2324           & 0.6650           \\ 
\cmidrule{3-8}
& & (PS)                   & Token Probs.            & 0.2316           & 0.0549           & 0.0631           & 0.6650          \\ 
\cmidrule{3-8}
& & \multirow{3}{*}{(MPS)} & \quad + Nucleus         & \textbf{0.2130}  & \textbf{0.0290}  & \textbf{0.0267}  & \textbf{0.7203} \\ 
& &                        & \quad + Beam            & 0.2215           & 0.0384           & 0.0416           & 0.6921          \\ 
& &                        & \quad + N + B           & 0.2141           & 0.0314           & 0.0323           & 0.7173          \\

\bottomrule
\end{tabular}
\caption{
We report how the performance of Multiplicative Platt Scaling (\textbf{MPS}) is affected by changes in whether the parsed frequencies are derived from nucleus sampling only (\textbf{Nucleus}), beam search only (\textbf{Beam}), and both nucleus sampling and beam search (\textbf{N + B}). For reference, we also include the performance of only using token probabilities (i.e., \textbf{PS}). We report Brier score, expected calibration error (ECE), and adaptive calibration Error (ACE) on the SPIDER and BIRD datasets. ECE and ACE are calculated using 10 bins. In addition, we report AUC for the task of error detection.
}
\label{tab:calibration+auc_scf}
\end{table*}


\paragraph{Stratified Magnitude Analysis.}

We study the magnitude of adding features derived from parsing additional samples and how this impacts calibration scores. We thus analyze outputs with the largest (absolute) difference in output probability between MPS and PS by aggregating the top and bottom $1\%$, $5\%$, $10\%$, and $20\%$ of samples, ordered by the difference ($\Delta$) in the calibrated probability produced by PS and MPS. 

We see that across all strata, MPS produces probabilities that are more calibrated, suggesting that queries for which MPS and PS \textit{disagree on the most} are those for which \textit{MPS is better calibrated on}. For example, in the top 1\%, PS on average produces a probability of $0.464$ while MPS produces an average probability of $0.825$, which is closer to the true proportion of correct samples of $0.714$. 
Full results are in the Table \ref{tab:prob_change}, where we show how and to what degree MPS shifts its calibrated probability output when compared to PS. In particular, we show the average probability for top $1\%$, $5\%$, $10\%$, and $20\%$ samples, sorted by the change $\Delta$ in probability between MPS and PS. We see that MPS shifts the probabilities so that for each set of examples, the calibration error is smaller when compared to PS. Finally, we show examples of queries with the largest absolute difference in probability between MPS and PS in the Tables \ref{tab:cal_ex_spider_t5}, \ref{tab:cal_ex_spider_llama_8b}, \ref{tab:cal_ex_bird_llama_8b} and \ref{tab:cal_ex_bird_llama_70b}. 

\begin{table*}[t!]
\centering
\begin{tabular}{c c c l c c c c}
\toprule
 \textbf{Dataset} & \textbf{Model} &  &  & \textbf{Prob. $\Delta$} & \textbf{PS} & \textbf{MPS} & \textbf{Correct} \\
\toprule

\multirow{16}{*}{SPIDER}

& \multirow{8}{*}{T5 3B} 
& \multirow{2}{*}{1\%}  & top    & + 0.361 & 0.464 & \textbf{0.825} & 0.714 \\
& &                     & bottom & - 0.380 & 0.739 & \textbf{0.359} & 0.476 \\
\cline{3- 8}
&
& \multirow{2}{*}{5\%}  & top    & + 0.223 & 0.635 & \textbf{0.857} & 0.764 \\
& &                     & bottom & - 0.280 & 0.680 & \textbf{0.400} & 0.425 \\
\cline{3- 8}
&
& \multirow{2}{*}{10\%} & top    & + 0.187 & 0.729 & \textbf{0.916} & 0.859 \\
& &                     & bottom & - 0.224 & 0.682 & \textbf{0.458} & 0.521 \\
\cline{3- 8}
&
& \multirow{2}{*}{20\%} & top    & + 0.169 & 0.775 & \textbf{0.944} & 0.904 \\
& &                     & bottom & - 0.167 & 0.692 & \textbf{0.525} & 0.528 \\

\cline{2- 8}

& \multirow{8}{*}{Llama 3.1 8B Instruct} 
& \multirow{2}{*}{1\%}  & top    & + 0.310 & 0.591 & \textbf{0.902} & 0.810 \\
& &                     & bottom & - 0.351 & 0.638 & \textbf{0.287} & 0.429 \\
\cline{3- 8}
&
& \multirow{2}{*}{5\%}  & top    & + 0.254 & 0.631 & \textbf{0.884} & 0.879 \\
& &                     & bottom & - 0.276 & 0.654 & \textbf{0.378} & 0.477 \\
\cline{3- 8}
&
& \multirow{2}{*}{10\%} & top    & + 0.227 & 0.651 & \textbf{0.878} & 0.888 \\
& &                     & bottom & - 0.234 & 0.659 & \textbf{0.425} & 0.500 \\
\cline{3- 8}
&
& \multirow{2}{*}{20\%} & top    & + 0.189 & 0.669 & \textbf{0.858} & 0.857 \\
& &                     & bottom & - 0.180 & 0.671 & \textbf{0.491} & 0.577 \\

\midrule

\multirow{16}{*}{BIRD}

& \multirow{8}{*}{Llama 3.1 8B Instruct} 
& \multirow{2}{*}{1\%}  & top    & + 0.371 & 0.386 & \textbf{0.757} & 0.929 \\
& &                     & bottom & - 0.236 & 0.480 & \textbf{0.244} & 0.286 \\
\cline{3- 8}
&
& \multirow{2}{*}{5\%}  & top    & + 0.305 & 0.400 & \textbf{0.705} & 0.676 \\
& &                     & bottom & - 0.200 & 0.409 & \textbf{0.210} & 0.149 \\
\cline{3- 8}
&
& \multirow{2}{*}{10\%} & top    & + 0.267 & 0.410 & \textbf{0.677} & 0.642 \\
& &                     & bottom & - 0.181 & 0.396 & \textbf{0.215} & 0.176 \\
\cline{3- 8}
&
& \multirow{2}{*}{20\%} & top    & + 0.211 & 0.402 & \textbf{0.613} & 0.598 \\
& &                     & bottom & - 0.156 & 0.378 & \textbf{0.222} & 0.209 \\

\cline{2- 8}

& \multirow{8}{*}{Llama 3.1 70B Instruct} 
& \multirow{2}{*}{1\%}  & top    & + 0.330 & 0.456 & \textbf{0.786} & 0.857 \\
& &                     & bottom & - 0.376 & 0.658 & \textbf{0.282} & 0.429 \\
\cline{3- 8}
&
& \multirow{2}{*}{5\%}  & top    & + 0.272 & 0.486 & \textbf{0.758} & 0.808 \\
& &                     & bottom & - 0.297 & 0.556 & \textbf{0.259} & 0.397 \\
\cline{3- 8}
&
& \multirow{2}{*}{10\%} & top    & + 0.243 & 0.508 & \textbf{0.751} & 0.774 \\
& &                     & bottom & - 0.261 & 0.519 & \textbf{0.258} & 0.356 \\
\cline{3- 8}
&
& \multirow{2}{*}{20\%} & top    & + 0.205 & 0.520 & \textbf{0.724} & 0.757 \\
& &                     & bottom & - 0.214 & 0.507 & \textbf{0.293} & 0.353 \\

\bottomrule

\end{tabular}
\caption{For each combination of dataset and model, we aggregate the top and bottom 1\%, 5\%, 10\%, and 20\% of data points, ordered by the difference ($\Delta$) in the calibrated probability ouputted by Platt scaling (PS) and multiplicative Platt scaling (MPS) to summarize how adding features derived from parsing additional samples affects calibration. In addition, we report the proportion of outputs in each quantile that are actually correct. Between PS and MPS, we \textbf{bold} the method who has the smaller calibration error in each bucket. As show in these results, among the outputs for which MPS and PS differ the most (i.e., each quantile), MPS produces more calibrated probabilities in all case.}
\label{tab:prob_change}
\end{table*}


\paragraph{Analysis by Query Complexity.}

\begin{table*}[htbp!]
\centering

\begin{tabular}{lllllrrrr}
\toprule
\textbf{Model} & \textbf{Difficulty} & \textbf{Size} & \textbf{Accuracy} & \textbf{Method} & \textbf{Brier} & \textbf{ECE} & \textbf{ACE} & \textbf{AUC} \\
\toprule
\multirow{8}{*}{T5 3B} & \multirow{2}{*}{easy} & \multirow{2}{*}{22\%} & \multirow{2}{*}{0.917} 
          & PS & 0.081 & 0.126 & 0.115 & 0.691 \\
 &  &  &  & MPS & 0.076 & 0.075 & 0.069 & 0.714 \\
\cline{2-9}
 & \multirow{2}{*}{medium} & \multirow{2}{*}{40\%} & \multirow{2}{*}{0.770} 
          & PS & 0.161 & 0.052 & 0.057 & 0.678 \\
 &  &  &  & MPS & 0.153 & 0.023 & 0.030 & 0.756 \\
\cline{2-9}
 & \multirow{2}{*}{hard} & \multirow{2}{*}{22\%} & \multirow{2}{*}{0.641} 
          & PS & 0.198 & 0.081 & 0.087 & 0.727 \\
 &  &  &  & MPS & 0.188 & 0.060 & 0.060 & 0.764 \\
\cline{2-9}
 & \multirow{2}{*}{extra} & \multirow{2}{*}{16\%} & \multirow{2}{*}{0.533} 
          & PS & 0.246 & 0.155 & 0.155 & 0.674 \\
 &  &  &  & MPS & 0.234 & 0.135 & 0.123 & 0.711 \\
 \midrule
\multirow{8}{*}{\textsc{Llama 3.1 8B Instruct}} & \multirow{2}{*}{easy} & \multirow{2}{*}{22\%} & \multirow{2}{*}{0.893} 
          & PS & 0.096 & 0.089 & 0.087 & 0.721 \\
 &  &  &  & MPS & 0.085 & 0.044 & 0.043 & 0.778 \\
\cline{2-9}
 & \multirow{2}{*}{medium} & \multirow{2}{*}{40\%} & \multirow{2}{*}{0.783} 
          & PS & 0.164 & 0.082 & 0.083 & 0.697 \\
 &  &  &  & MPS & 0.151 & 0.051 & 0.049 & 0.741 \\
\cline{2-9}
 & \multirow{2}{*}{hard} & \multirow{2}{*}{22\%} & \multirow{2}{*}{0.646} 
          & PS & 0.220 & 0.035 & 0.053 & 0.628 \\
 &  &  &  & MPS & 0.221 & 0.071 & 0.068 & 0.628 \\
\cline{2-9}
 & \multirow{2}{*}{extra} & \multirow{2}{*}{16\%} & \multirow{2}{*}{0.594} 
          & PS & 0.233 & 0.063 & 0.103 & 0.587 \\
 &  &  &  & MPS & 0.219 & 0.035 & 0.037 & 0.679 \\
\bottomrule
\end{tabular}

\caption{
Results on SPIDER, grouped by the difficulty of the SQL query. We observe that ECE, ACE, and AUC (for error detection) are not well-correlated with the difficulty of the query. Note that while for completeness, we report Brier score, it cannot be use to compare between levels of difficulty since the model accuracy for each difficulty varies substantially (See Section \ref{sec:analysis_compare_models}) for a more detailed explanation.
}
\label{tab:calibration+auc_difficulty_spider}
\end{table*}

\begin{table*}[htbp!]
\centering

\begin{tabular}{lllllrrrr}
\toprule
\textbf{Model} & \textbf{Difficulty} & \textbf{Size} & \textbf{Accuracy} & \textbf{Method} & \textbf{Brier} & \textbf{ECE} & \textbf{ACE} & \textbf{AUC} \\
\toprule
\multirow{6}{*}{\textsc{Llama 3.1 8B Instruct}} & \multirow{2}{*}{simple} & \multirow{2}{*}{62\%} & \multirow{2}{*}{0.420} 
          & PS & 0.226 & 0.044 & 0.076 & 0.681 \\
 &  &  &  & MPS & 0.209 & 0.043 & 0.043 & 0.720 \\
\cline{2-9}
 & \multirow{2}{*}{moderate} & \multirow{2}{*}{29\%} & \multirow{2}{*}{0.286} 
          & PS & 0.201 & 0.066 & 0.070 & 0.621 \\
 &  &  &  & MPS & 0.185 & 0.064 & 0.055 & 0.708 \\
\cline{2-9}
 & \multirow{2}{*}{challenging} & \multirow{2}{*}{9\%} & \multirow{2}{*}{0.173} 
          & PS & 0.161 & 0.163 & 0.230 & 0.778 \\
 &  &  &  & MPS & 0.135 & 0.147 & 0.137 & 0.801 \\
 \midrule
 \multirow{6}{*}{\textsc{Llama 3.1 70B Instruct}} & \multirow{2}{*}{simple} & \multirow{2}{*}{62\%} & \multirow{2}{*}{0.591} 
          & PS & 0.228 & 0.074 & 0.076 & 0.655 \\
 &  &  &  & MPS & 0.210 & 0.034 & 0.041 & 0.714 \\
\cline{2-9}
 & \multirow{2}{*}{moderate} & \multirow{2}{*}{29\%} & \multirow{2}{*}{0.479} 
          & PS & 0.239 & 0.040 & 0.039 & 0.627 \\
 &  &  &  & MPS & 0.222 & 0.045 & 0.036 & 0.688 \\
\cline{2-9}
 & \multirow{2}{*}{challenging} & \multirow{2}{*}{9\%} & \multirow{2}{*}{0.324}
          & PS & 0.230 & 0.125 & 0.119 & 0.581 \\
 &  &  &  & MPS & 0.213 & 0.104 & 0.120 & 0.632 \\
\bottomrule
\end{tabular}

\caption{
Results on BIRD, grouped by the difficulty of the SQL query. We observe that ECE, ACE, and AUC (for error detection) are not well-correlated with the difficulty of the query. Note that while for completeness, we report Brier score, it cannot be use to compare between levels of difficulty since the model accuracy for each difficulty varies substantially (See Section \ref{sec:analysis_compare_models}) for a more detailed explanation.
}
\label{tab:calibration+auc_difficulty_bird}
\end{table*}

In the SPIDER dataset, each example is annotated with difficulty ratings of \textit{easy, medium, hard,} and \textit{extra hard}~\citet{yu2018spider}, while in BIRD, examples are annotated with difficult ratings of \textit{simple, moderate,} and \textit{challenging}~\citet{li2024can}. We calculate metrics for each level of difficulty in order to see if there is any relationship between query difficulty and calibration, as identified in \citet[Figure 6]{stengel2023calibrated} for LLMs' uncalibrated probability outputs. However, our experiments---unlike previous work---do not show consistent patterns in ECE, ACE and AUC across models and datasets if performance is measured after post-hoc calibration (i.e., applying PS and MPS).

Note that while for completeness, we report Brier score in Tables \ref{tab:calibration+auc_difficulty_spider} and \ref{tab:calibration+auc_difficulty_bird}, we cannot use Brier score to compare between levels of difficulty since the model accuracy for each difficulty varies substantially (See Section \ref{sec:analysis_compare_models}) for a more detailed explanation. 


\paragraph{Weight / Sub-clause Analysis of MPS.}

We conduct additional analysis on the weights produced for MPS on outputs from \textsc{Llama 3.1 8B Instruct}.
In Tables \ref{tab:weight_analysis_spider} and \ref{tab:weight_analysis_bird}, we report the sum and absolute sum of (standardized) weights corresponding to each sub-clause type for SPIDER and BIRD respectively. Generally, the learned feature weights represent to what extent the final probability of correctness should be shifted, given how consistent the language model is in generating a particular sub-clause. In addition, conduct an ablation study in which we remove features corresponding to each sub-clause type as input into MPS. Tables \ref{tab:subclause_ablation_spider} and \ref{tab:subclause_ablation_bird} show the impact of each sub-clause on the calibration metrics for SPIDER and BIRD respectively.

We find that our aggregate sub-clause score (\texttt{AGG}) is the most significant feature utilized by MPS, which is expected given that it aggregates information from all sub-clauses (i.e., product of SCF scores). Not only does it have the highest weight (Tables \ref{tab:weight_analysis_spider} and \ref{tab:weight_analysis_bird}), but removing it from MPS most significantly degrades ACE and ECE for both SPIDER and BIRD (Tables \ref{tab:subclause_ablation_spider} and \ref{tab:subclause_ablation_bird}). On the other hand, we find that the SCF scores corresponding to \texttt{DISTINCT}, \texttt{HAVING}, \texttt{LIMIT}, and set operations have small weights (Tables \ref{tab:weight_analysis_spider} and \ref{tab:weight_analysis_bird}). Removing these clauses also has minimal impact on MPS's performance (Tables \ref{tab:subclause_ablation_spider} and \ref{tab:subclause_ablation_bird}).
While trends for the remaining sub-clauses are not as consistent between the two datasets, they are still unsurprising. For example, removing \texttt{SELECT} and \texttt{WHERE} features, two of the more important and common sub-clauses for SQL, significantly impacts MPS for BIRD, especially with respect to AUC. Finally, we note that while the improvement of MPS over PS with respect to Brier score illustrates that incorporating all sub-clauses as a whole is important for calibration (i.e., our main empirical result), Tables \ref{tab:subclause_ablation_spider} and \ref{tab:subclause_ablation_bird} show that removing any individual sub-clause does not significantly impact Brier score for MPS.

\begin{table}[ht!]
\centering
\begin{tabular}{l | c c}
\toprule
\textbf{sub-clause} & \textbf{abs. weight ($\downarrow$)} & \textbf{weight} \\
\toprule

\texttt{FROM}     & 0.826 	& 0.227 \\
\texttt{AGG} 	 & 0.687 	& 0.687 \\
\texttt{SELECT} 	 & 0.507 	& -0.137 \\
\texttt{ORDER}	 & 0.282 	& 0.134 \\
\texttt{WHERE} 	 & 0.260 	& 0.082 \\
\texttt{GROUP} 	 & 0.199 	& 0.199 \\
\texttt{ON} 	     & 0.191 	& -0.186 \\
\texttt{HAVING} 	 & 0.168 	& 0.129 \\
\texttt{DISTINCT} & 0.131 	& 0.131 \\
\texttt{LIMIT} 	 & 0.102 	& 0.102 \\
\texttt{Set Op.}  & 0.038 	& 0.038  \\

\bottomrule
\end{tabular}
\caption{
We report the sum and absolute sum of (standardized) weights corresponding to each sub-clause type for MPS trained on \textsc{Llama 3.1 8B Instruct} outputs on SPIDER.
}
\label{tab:weight_analysis_spider}
\end{table}

\begin{table}[ht!]
\centering
\begin{tabular}{l | c c}
\toprule
\textbf{sub-clause} & \textbf{abs. weight ($\downarrow$)} & \textbf{weight} \\
\toprule

\texttt{AGG}       & 0.343 & 0.343 \\
\texttt{SELECT}    & 0.310 & 0.302 \\
\texttt{WHERE}     & 0.255 & 0.242 \\
\texttt{ORDER}     & 0.224 & 0.087 \\
\texttt{ON}        & 0.205 & -0.158 \\
\texttt{FROM}      & 0.180 & 0.172 \\
\texttt{GROUP}     & 0.178 & 0.165 \\
\texttt{DISTINCT}  & 0.081 & -0.010 \\
\texttt{HAVING}    & 0.077 & 0.041 \\
\texttt{LIMIT}     & 0.066 & -0.022 \\
\texttt{Set Op.}   & 0.029 & 0.016 \\

\bottomrule
\end{tabular}
\caption{
We report the sum and absolute sum of (standardized) weights corresponding to each sub-clause type for MPS trained on \textsc{Llama 3.1 8B Instruct} outputs on BIRD.
}
\label{tab:weight_analysis_bird}
\end{table}

\begin{table}[ht!]
\centering
\begin{tabular}{l | c c c c}
\toprule

\textbf{Removed Feature} & \textbf{Brier} & \textbf{ECE} & \textbf{ACE} & \textbf{AUC} \\

\toprule

---    & 0.164 & 0.023 & 0.023 & 0.746 \\
\midrule
\textbf{\texttt{AGG}}      & 0.001 & \textbf{0.006} & \textbf{0.008} & 0.000 \\
\texttt{Set Op.}  & 0.000 & 0.000 & 0.000 & -0.000 \\
\texttt{DISTINCT} & -0.000 & -0.002 & 0.004 & 0.001 \\
\textbf{\texttt{SELECT}}   & -0.001 & -0.001 & \textbf{0.005} & 0.002 \\
\textbf{\texttt{FROM}}     & 0.001 & 0.002 & 0.002 & \textbf{-0.005} \\
\texttt{ON}       & 0.000 & 0.000 & 0.000 & 0.001 \\
\textbf{\texttt{WHERE}}    & -0.000 & \textbf{0.005} & 0.002 & 0.002 \\
\textbf{\texttt{GROUP}}    & -0.001 & \textbf{0.006} & \textbf{0.008} & 0.004 \\
\texttt{HAVING}   & 0.000 & 0.000 & 0.000 & -0.001 \\
\texttt{ORDER}    & 0.000 & -0.000 & -0.000 & -0.000 \\
\texttt{LIMIT}    & -0.000 & 0.001 & 0.001 & 0.000 \\

\bottomrule
\end{tabular}
\caption{
We report the change in performance of MPS when sub-clauses are removed as input signals. The ablation study is conducted on MPS trained using \textsc{Llama 3.1 8B Instruct} outputs for SPIDER. We bold values that are significantly large (delta greater than 0.005).
}
\label{tab:subclause_ablation_spider}
\end{table}

\begin{table}[ht!]
\centering
\begin{tabular}{l | c c c c}
\toprule

\textbf{Removed Feature} & \textbf{Brier} & \textbf{ECE} & \textbf{ACE} & \textbf{AUC} \\

\toprule

---   & 0.194 & 0.035 & 0.032 & 0.737 \\
\midrule
\textbf{\texttt{AGG}}      & 0.001 & \textbf{0.016} & \textbf{0.015} & -0.003 \\
\texttt{Set Op.}  & -0.000 & -0.000 & -0.000 & 0.000 \\
\texttt{DISTINCT} & -0.000 & 0.002 & 0.001 & 0.001 \\
\textbf{\texttt{SELECT}}   & \textbf{0.005} & 0.002 & \textbf{0.008} & \textbf{-0.017} \\
\texttt{FROM}     & 0.000 & -0.001 & 0.003 & 0.000 \\
\textbf{\texttt{ON}}       & 0.001 & \textbf{0.005} & 0.000 & -0.003 \\
\textbf{\texttt{WHERE}}    & 0.002 & -0.000 & \textbf{0.008} & \textbf{-0.007} \\
\textbf{\texttt{GROUP}}    & -0.000 & \textbf{0.005} & -0.000 & 0.000 \\
\texttt{HAVING}   & -0.000 & 0.002 & 0.000 & 0.000 \\
\textbf{\texttt{ORDER}}    & 0.000 & \textbf{0.006} & \textbf{0.005} & 0.001 \\
\texttt{LIMIT}    & 0.000 & 0.001 & 0.001 & -0.001 \\

\bottomrule
\end{tabular}
\caption{
We report the change in performance of MPS when sub-clauses are removed as input signals. The ablation study is conducted on MPS trained using \textsc{Llama 3.1 8B Instruct} outputs for BIRD. We bold values that are significantly large (delta greater than 0.005).
}
\label{tab:subclause_ablation_bird}
\end{table}



\paragraph{Examples of when MPS outperforms PS the most.}

In Tables \ref{tab:cal_ex_spider_t5}, \ref{tab:cal_ex_spider_llama_8b}, \ref{tab:cal_ex_bird_llama_8b}, \ref{tab:cal_ex_bird_llama_70b}, we show specific examples of queries from the SPIDER and BIRD datasets for which MPS and PS differ the most. In particular, we show examples of correct outputs where MPS shifts the calibrated probability upwards the most and incorrect outputs where MPS shifts probabilities downwards. 

\begin{table*}[htbp!]
\centering
\begin{tabular}{p{1.1in} p{2.45in} c c c c c}
\toprule
\textbf{Question} & \textbf{Output} & \textbf{Label} &\textbf{PS} & \textbf{MPS} & \textbf{$\Delta$} \\
\toprule

Return the names and ages of pilots who have flown Piper Cub and are older than 35, or have flown the F-14 Fighter and are younger than 30. 
& \vspace{-1.2em} \begin{lstlisting}[style=correct]
SELECT pilot_name,
       age 
FROM   pilotskills
WHERE  plane_name='Piper Cub' 
       AND age > 35
UNION
SELECT pilot_name,
       age  
FROM   pilotskills
WHERE  plane_name='F-14 Fighter'
       AND age < 30
\end{lstlisting} 
& Correct & 0.275 & 0.928 & \textbf{+ 0.652} \\


Find the receipt numbers where both Cake and Cookie were bought.
& \vspace{-1.2em} \begin{lstlisting}[style=incorrect]
SELECT t1.receiptnumber
FROM   receipts AS t1
       JOIN goods AS t2
         ON t1.customerid=t2.id
WHERE  t2.food="Cake"
INTERSECT
SELECT t1.receiptnumber
FROM   receipts AS t1
       JOIN goods AS t2
         ON t1.customerid=t2.id
WHERE  t2.food="Cookie" 
\end{lstlisting} 
& Incorrect & 0.611 & 0.145 & \textbf{- 0.466} \\


\bottomrule

\end{tabular}
\caption{Example SQL outputs from \textbf{\textsc{T5 3B}} on the \textbf{BIRD} dataset, which were selected in the following way: We calculate the difference ($\Delta)$ in (calibrated) probability outputted by Platt scaling (PS, \textit{baseline}) and multiplicative Platt scaling (MPS, \textit{ours}). We then split outputs by their label (whether the output is \colorbox{ForestGreen!15}{correct} or \colorbox{red!15}{incorrect}). For correct examples, we select the one with the largest positive change in $\Delta$, and for incorrect examples, we select the one with the largest negative change in $\Delta$.}
\label{tab:cal_ex_spider_t5}
\end{table*}
\begin{table*}[htbp!]
\centering
\begin{tabular}{p{1.1in} p{2.45in} c c c c c}
\toprule
\textbf{Question} & \textbf{Output} & \textbf{Label} &\textbf{PS} & \textbf{MPS} & \textbf{$\Delta$} \\
\toprule

What is the top speed and power of the vehicle manufactured in the year of 1996? 
& \vspace{-1.2em} \begin{lstlisting}[style=correct]
SELECT top_speed,
       power
FROM   vehicle
WHERE  build_year=1996 
\end{lstlisting} 
& Correct & 0.551 & 0.942 & \textbf{+ 0.391} \\


Find the name of all the services which either have been used by customer "Hardy Kutch" or have been rated as "good" in one of the customer interactions.
& \vspace{-1.2em} \begin{lstlisting}[style=incorrect]
SELECT T1.service_details
FROM   services AS T1
INNER JOIN
  customers_and_services AS T2
ON T1.service_id=T2.service_id
INNER JOIN
  customer_interactions AS T3
ON T2.customer_id=T3.customer_id
WHERE T3.status_code='good'
OR T2.customer_id=(
SELECT customer_id
FROM   customers
WHERE 
  customer_details='Hardy Kutch'
) 
\end{lstlisting} 
& Incorrect & 0.654 & 0.237 & \textbf{- 0.417} \\


\bottomrule

\end{tabular}
\caption{Example SQL outputs from \textbf{\textsc{Llama 3.1 8B Instruct}} on the \textbf{SPIDER} dataset, which were selected in the following way: We calculate the difference ($\Delta)$ in (calibrated) probability outputted by Platt scaling (PS, \textit{baseline}) and multiplicative Platt scaling (MPS, \textit{ours}). We then split outputs by their label (whether the output is \colorbox{ForestGreen!15}{correct} or \colorbox{red!15}{incorrect}). For correct examples, we select the one with the largest positive change in $\Delta$, and for incorrect examples, we select the one with the largest negative change in $\Delta$.}
\label{tab:cal_ex_spider_llama_8b}
\end{table*}

\begin{table*}[htbp!]
\centering
\begin{tabular}{p{1.1in} p{2.45in} c c c c c}
\toprule
\textbf{Question} & \textbf{Output} & \textbf{Label} &\textbf{PS} & \textbf{MPS} & \textbf{$\Delta$} \\
\toprule

How many majors are there in "College of Humanities and Social Sciences"?
& \vspace{-1.2em} \begin{lstlisting}[style=correct]
SELECT COUNT(*)
FROM   major
WHERE  college='college of 
                humanities and 
                social sciences' 
\end{lstlisting} 
& Correct & 0.377 & 0.783 & \textbf{+ 0.406} \\


Which different county has the most number of closed schools? Please provide the name of each school as well as the closure date.
& \vspace{-1.2em} \begin{lstlisting}[style=incorrect]
SELECT county,
       school,
       closeddate
FROM   schools
WHERE  statustype='Closed'
\end{lstlisting} 
& Incorrect & 0.461 & 0.186 & \textbf{- 0.274} \\


\bottomrule

\end{tabular}
\caption{Example SQL outputs from \textbf{\textsc{Llama 3.1 8B Instruct}} on the \textbf{BIRD} dataset, which were selected in the following way: We calculate the difference ($\Delta)$ in (calibrated) probability outputted by Platt scaling (PS, \textit{baseline}) and multiplicative Platt scaling (MPS, \textit{ours}). We then split outputs by their label (whether the output is \colorbox{ForestGreen!15}{correct} or \colorbox{red!15}{incorrect}). For correct examples, we select the one with the largest positive change in $\Delta$, and for incorrect examples, we select the one with the largest negative change in $\Delta$.}
\label{tab:cal_ex_bird_llama_8b}
\end{table*}
\begin{table*}[htbp!]
\centering
\begin{tabular}{p{1.1in} p{2.45in} c c c c c}
\toprule
\textbf{Question} & \textbf{Output} & \textbf{Label} &\textbf{PS} & \textbf{MPS} & \textbf{$\Delta$} \\
\toprule

What segment did the customer have at 2012/8/23 21:20:00?
& \vspace{-1.2em} \begin{lstlisting}[style=correct]
SELECT T1.segment
FROM   customers AS T1
INNER JOIN transactions_1k AS T2
ON T1.customerid=T2.customerid
WHERE  T2.date='2012-08-23'
   AND T2.time='21:20:00' 
\end{lstlisting} 
& Correct & 0.443 & 0.804 & \textbf{+ 0.361} \\


What is the title of the post that is owned by csgillespie and has the highest popularity?
& \vspace{-1.2em} \begin{lstlisting}[style=incorrect]
SELECT title
FROM   posts
WHERE 
  ownerdisplayname='csgillespie'
ORDER BY viewcount DESC
LIMIT 1 
\end{lstlisting} 
& Incorrect & 0.686 & 0.256 & \textbf{- 0.429} \\


\bottomrule

\end{tabular}
\caption{Example SQL outputs from \textbf{\textsc{Llama 3.1 70B Instruct}} on the \textbf{BIRD} dataset, which were selected in the following way: We calculate the difference ($\Delta)$ in (calibrated) probability outputted by Platt scaling (PS, \textit{baseline}) and multiplicative Platt scaling (MPS, \textit{ours}). We then split outputs by their label (whether the output is \colorbox{ForestGreen!15}{correct} or \colorbox{red!15}{incorrect}). For correct examples, we select the one with the largest positive change in $\Delta$, and for incorrect examples, we select the one with the largest negative change in $\Delta$.}
\label{tab:cal_ex_bird_llama_70b}
\end{table*}

\clearpage

\section{Deriving parsed frequencies (SCF)}\label{appx:additional_scf}

In Figure \ref{fig:parsing_example} of the main body, we provide an example of how we parse a SQL query. As noted in the figure, however, we make two simplifications in the example: (1) We assume the SQL query is composed of a single SELECT statement and (2) we only count sub-clauses for \texttt{SELECT}, \texttt{FROM}, \texttt{ON}, \texttt{WHERE}, and \texttt{GROUP BY}. We now provide the exact details of our parsing procedure. 

\begin{figure}[h!]
    \centering
    \includegraphics[width=1.0\columnwidth]{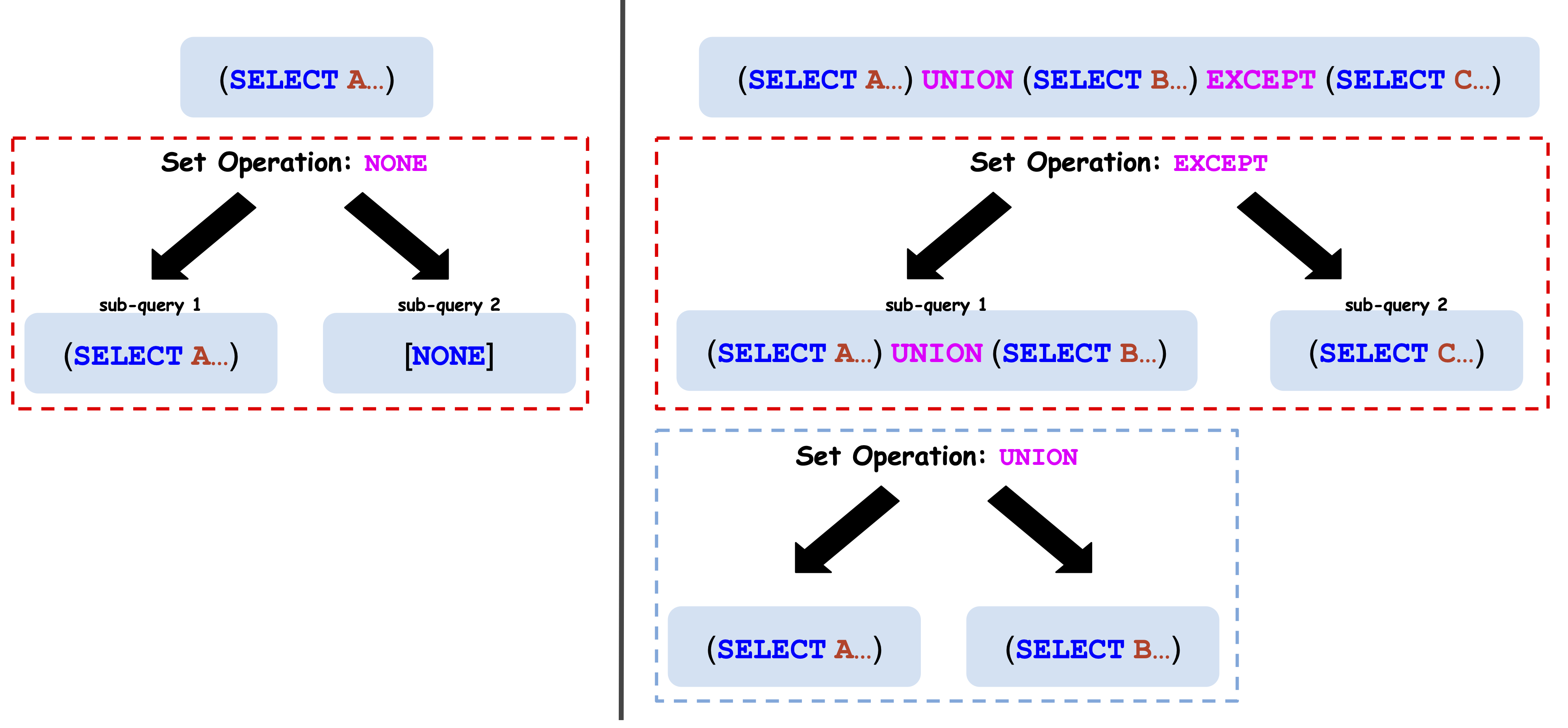}
    \caption{
        We illustrate two examples on how a SQL query is parsed via its tree structure representation. The red box denotes the set operation and two sub-queries that are used for calculating our SCF signals. In the \textbf{left} example, we have a SQL statement that is \textbf{not} composed of multiple \texttt{SELECT} statements. Hence, the set operation is denoted as NONE, and we derive two sub-queries, one of which is the original statement itself (\texttt{SELECT A...}) and the other of which is simply [NONE]. In the \textbf{right} example, we parse it via the logicical execution order of the query (done automatically via standard SQL parsing libraries), giving us a set operation \texttt{EXCEPT} with two sub-queries. We note sub-query 1 can be further parsed into an additional tree (blue box) with the set operation \texttt{UNION}.
    }
    \label{fig:sql_parse_tree}
\end{figure}

\paragraph{Queries composed of multiple \texttt{SELECT} statements.} 

In our dataset, (two) \texttt{SELECT} statements can be composed together using the set operators \texttt{UNION}, \texttt{INTERSECT}, and \texttt{EXCEPT}. In these cases, we then individually parse each \texttt{SELECT} statement that is being composed together using the set operator (i.e., if we represent the SQL statement in its tree structure, we then traverse the tree). Note that theoretically, the sub-statements being composed together may be another query that is composed (via another set operator) of multiple sub-statements. Thus, we can recursively traverse the tree structure of the sub-statements until we reach a ``leaf'', which in our case, is a SQL statement that does not contain any set operators (i.e., a single SELECT statement). We illustrate tree traversal in Figure \ref{fig:sql_parse_tree}.

\paragraph{Additional sub-clauses not shown in Figure~\ref{fig:parsing_example}.}

We parse each (single) \texttt{SELECT} statement (found at leaves of a SQL output's tree representation (Figure \ref{fig:sql_parse_tree})) into the following nine sub-clauses / keywords:
\begin{itemize}[noitemsep]
    \item \texttt{DISTINCT}
    \item \texttt{SELECT}
    \item \texttt{FROM}
    \item \texttt{ON}
    \item \texttt{WHERE}
    \item \texttt{GROUP BY}
    \item \texttt{HAVING}
    \item \texttt{ORDER BY}
    \item \texttt{LIMIT}    
\end{itemize}
If a query does not contain one of the above, we tag it as \textbf{\texttt{NONE}}. Parsing was done automatically using the SQLGlot library (\href{https://github.com/tobymao/sqlglot}{link}).

\paragraph{Computing the final SCF scores.}\label{appx:sql_parse_tree}

\begin{algorithm}[ht!]
\caption{\textsf{SCF} (sub-clause frequency scorer)}
\begin{algorithmic}[1]

\STATE \textbf{Input}: SQL query $Q$ and set of additional queries $\mathbf{Q}_\textrm{additional}$
\RETURN $\frac{1}{|\mathbf{Q}_\textrm{additional}|}\sum_{Q' \in \mathbf{Q}_\textrm{additional}} \textsf{Q-MATCH}(Q, Q')$

\end{algorithmic}
\label{alg:scf}
\end{algorithm}


\begin{algorithm}[ht!]
\caption{\textsf{Q-MATCH} (query match function)}
\begin{algorithmic}[1]

\STATE \textbf{Input}: SQL queries $Q_A$ and $Q_B$
\STATE Let $Q_{A_1}$ and $Q_{A_2}$ be the subqueries of $Q_A$
\COMMENT{See Figure~\ref{fig:sql_parse_tree}; red box}
\STATE Let $Q_{B_1}$ and $Q_{B_2}$ be the subqueries of $Q_B$

\STATE Set $s_{\textrm{set\_op}} = \mathbbm{1} \left[ \textsf{SET\_OP}(Q_A) = \textsf{SET\_OP}(Q_B) \right]$
\\ \COMMENT{SET\_OP returns the set operation (e.g., \texttt{UNION}) of the input query}
\STATE Set $\mathbf{s}_{1} = \left[ \textsf{SQ-MATCH}(Q_{A_1}, Q_{B_1}), \textsf{SQ-MATCH}(Q_{A_2}, Q_{B_2}) \right]$
\COMMENT{Algorithm~\ref{alg:sq_match}}
\STATE Set $\mathbf{s}_{2} = \left[ \textsf{SQ-MATCH}(Q_{A_1}, Q_{B_2}), \textsf{SQ-MATCH}(Q_{A_2}, Q_{B_1}) \right]$

\IF{$\sum \mathbf{s}_{1} \ge \sum \mathbf{s}_{1}$}
    \RETURN $[s_{\textrm{set\_op}}, \mathbf{s}_{1}]$
\ELSE
    \RETURN $[s_{\textrm{set\_op}}, \mathbf{s}_{2}]$
\ENDIF

\end{algorithmic}
\label{alg:q_match}
\end{algorithm}


\begin{algorithm}[ht!]
\caption{\textsf{SQ-MATCH} (sub-query match function)}
\begin{algorithmic}[1]

\STATE \textbf{Input}: SQL (sub-)queries $Q_1$ and $Q_2$
\STATE Let the set of sub-clauses $\mathbf{C} =$ \{ \texttt{DISTINCT, SELECT, FROM, ON, WHERE, GROUP BY, HAVING, ORDER BY, LIMIT} \}

\FOR{\textrm{sub-clause} $c \in \mathbf{C}$}
    \IF{$Q_1 = \texttt{NONE}$ AND $Q_2 = \texttt{NONE}$}
        \STATE Set $s_{c} = 1$
    \ELSIF{$Q_1 = \texttt{NONE}$ AND $Q_2 \ne \texttt{NONE}$}
        \STATE Set $s_{c} = 0$
    \ELSIF{$Q_1 \ne \texttt{NONE}$ AND $Q_2 = \texttt{NONE}$}
        \STATE Set $s_{c} = 0$
    \ELSE
        \STATE Set $s_{c} = \textsf{TRAVERSE\_AND\_MATCH}(c, Q_1, Q_2)$ 
        \\\COMMENT{$\textsf{TRAVERSE\_AND\_MATCH}(c, Q_1, Q_2)$ traverses the tree representation (Figure~\ref{fig:sql_parse_tree}; blue box) of $Q_1$ and $Q_2$, returning $1$ (TRUE) if sub-clause $c$ matches at every corresponding node and $0$ (FALSE) otherwise.}
    \ENDIF
    
    \RETURN $(s_{c})_{c \in \mathbf{C}}$ \COMMENT{Return array of scores for each sub-clause}
\ENDFOR

\end{algorithmic}
\label{alg:sq_match}
\end{algorithm}

We summarize our procedure for calculating SCF scores in the pseudocode found in Algorithms \ref{alg:scf}, \ref{alg:q_match}, and \ref{alg:sq_match}. We also provide more details below:

To compute our sub-clause frequency scores, we must match the sub-clauses of each output to the set of additional outputs generated by the model. To constrain the number of signals $s_i$ in our framework, we only conduct matching (or frequency scoring) at the second level (starting from the root) of the tree (corresponding to the dotted red square in Figure \ref{fig:sql_parse_tree}), which we denote as sub-queries 1 and 2. This corresponds to checking for the proportion of exact matches on
\begin{itemize}[noitemsep]
    \item (1) the set operation (i.e., NONE, \texttt{UNION}, \texttt{INTERSECT}, \texttt{EXCEPT}) matches
    \item (9) sub-clauses for sub-query 1
    \item (9) sub-clauses for sub-query 2
\end{itemize}
In addition, we add a signal that is the product of the above $19$ scores as an additional feature.\footnote{We found in initial testing that adding this product of scores as an additional feature made results across experiments more consistent. However, we did not further investigate or tune this signal to optimize calibration performance.}
Thus in total, we have $20$ SCF signals $s_i$, in addition to the log sum token probability produced by the model directly (i.e., the one used for standard Platt scaling). Note that when using samples from nucleus sampling only or beam search only, we use these $20$ (+1) signals. In the case where use both (N + B), we concatenate these signals, giving us $40$ (+1) in total.

\paragraph{Matching sub-queries.}

We note some special cases when comparing two sub-queries, $Q_1$ and $Q_2$. These cases can also be found in Algorithm \ref{alg:sq_match}.

Cases:
\begin{itemize}[noitemsep]
    \item $Q_1 = \texttt{NONE}$ AND $Q_2 = \texttt{NONE}$:
    \begin{itemize}
        \item return TRUE for all sub-clauses
    \end{itemize}
    
    \item $Q_1 = \texttt{NONE}$ AND $Q_2 \ne \texttt{NONE}$:
    \begin{itemize}
        \item return FALSE for all sub-clauses
    \end{itemize}
    
    \item $Q_1 \ne \texttt{NONE}$ AND $Q_2 = \texttt{NONE}$:
    \begin{itemize}
        \item return FALSE for all sub-clauses
    \end{itemize}
    
    \item $Q_1 \ne \texttt{NONE}$ AND $Q_2 \ne \texttt{NONE}$:
    \begin{itemize}
        \item traverse the tree representation of subqueries $Q_1$ and $Q_2$ (Figure \ref{fig:sql_parse_tree}; blue box), return whether sub-clauses match at every node for each (of the $9$) sub-clause types.
    \end{itemize}
    
\end{itemize}

Finally, note that as discussed above, each query has two sub-queries $Q_{A_1}$, $Q_{A_2}$, $Q_{B_1}$, and $Q_{B_2}$. To determine which sets of sub-queries from $Q_A$ and $Q_B$ should go together, we compute the number of matches (over the $9$ sub-clauses) for all possible pairings
\begin{itemize}[noitemsep]
    \item $(Q_{A_1}, Q_{B_1})$ and $(Q_{A_2}, Q_{B_2})$
    \item $(Q_{A_1}, Q_{B_2})$ and $(Q_{A_2}, Q_{B_1})$
\end{itemize}
and choose the set of pairings that yield the highest number of exact sub-clause matches (out of $9 + 9 = 18$ total). This part of the procedure is described in Algorithm \ref{alg:q_match}.

\clearpage

\section{Additional experimental details}\label{appx:additional_exp_details}

\subsection{Prompts}

In our experiments, we used the following prompts described below. Note that for both prompts, the \textbf{\texttt{[schema]}} is serialized in the format: 
\textbf{\texttt{
[table] : [column] , [column], ... | [table] : ... | ...
}}

\subsubsection{SPIDER}

\paragraph{\textsc{T5 3B}.} The prompt format is as follows:
\textbf{\texttt{
[question] | [db\_id] | [schema]
}}

\paragraph{\textsc{Llama 3.1 8B Instruct}.} Using zero-shot prompting, the prompt format is as follows:
\\ \  \\
\noindent
\texttt{
\textbf{ \textless|start\_header\_id|\textgreater\ system \textless|end\_header\_id|\textgreater\ }
\newline
You are a helpful assistant who answers questions about database tables by responding with SQL queries. Users will provide you with a database id [DB\_ID], followed by the schema of the database.
\newline
The schema given by the user will be formatted as the following: TABLE\_1 : column\_1, column\_2, ... | TABLE\_2 : column\_1, column\_2, ... | ...
\newline
After, the user will ask a question. You should respond with a SQL query that can be executed on the provided database schema to answer the user's question.
\newline
Your response should be formatted as: [DB\_ID] | [SQL\_QUERY]. Your SQL query [SQL\_QUERY] should begin with SELECT and end with a semicolon.
\newline
\textbf{ \textless|start\_header\_id|\textgreater\ user \textless|end\_header\_id|\textgreater\
\newline [question] | [db\_id] | [schema] \newline
\textless|start\_header\_id|\textgreater\ assistant \textless|end\_header\_id|\textgreater\ }
}

\subsubsection{BIRD}

\paragraph{\textsc{Llama 3.1 8B Instruct / 70B Instruct}. } Using zero-shot prompting, the prompt format is as follows:
\\ \  \\
\noindent
\texttt{
\textbf{ \textless|start\_header\_id|\textgreater\ system \textless|end\_header\_id|\textgreater\ }
\newline
You are a helpful assistant who answers questions about database tables by responding with SQL queries. Users will provide you with a database id [DB\_ID], followed by the schema of the database.
\newline
The schema given by the user will be formatted as the following: TABLE\_1 : column\_1, column\_2, ... | TABLE\_2 : column\_1, column\_2, ... | ...
\newline
After, the user will ask a question. You should respond with a SQL query that can be executed on the provided database schema to answer the user's question.
\newline
Finally, users will provide additional evidence relating the question to the schema. You may use this information to help you answer the question
\newline
Your response should be formatted as: [DB\_ID] | [SQL\_QUERY]. Your SQL query [SQL\_QUERY] should begin with SELECT and end with a semicolon.
\newline
\textbf{ \textless|start\_header\_id|\textgreater\ user \textless|end\_header\_id|\textgreater\
\newline [question] | [db\_id] | [schema] | [evidence] \newline
\textless|start\_header\_id|\textgreater\ assistant \textless|end\_header\_id|\textgreater\ }
}

\subsection{Hyperparameters.} 

For generating SQL outputs using the models we evaluate, we use the Hugging Face's transformers library. To generate outputs using nucleus sampling, we set \texttt{top\_p=0.95}, \texttt{temperature=1.0}, and \texttt{num\_return\_sequences=10}. Using beam search, we set \texttt{num\_return\_sequences=10} and \texttt{num\_beams=20}. We set \texttt{max\_new\_tokens=512}.

For implementing Platt scaling and multivariate Platt scaling, we train a logistic regression model using \href{https://scikit-learn.org/1.5/modules/generated/sklearn.linear_model.LogisticRegression.html}{scikit-learn} with default parameters. We do not conduct any hyperparameter tuning when running these calibration methods.

\subsection{GPU requirements.}

For generating SQL outputs using \textsc{T5 3B PICARD} and \textsc{Llama 3.1 8B Instruct}, we use a single NVIDIA A100 80GB GPU. For generating outputs using \textsc{Llama 3.1 8B Instruct}, we use eight NVIDIA A100 80GB GPUs. Generations (for the calibration and test sets combined) took approximately the following number of GPU hours:
\begin{itemize}[noitemsep]
    \item SPIDER; \textsc{T5 3B PICARD} : 4 hours
    \item SPIDER; \textsc{LLama 3.1 8B Instruct} : 4 hours 
    \item BIRD; \textsc{LLama 3.1 8B Instruct} : 4 hours 
    \item BIRD; \textsc{LLama 3.1 70B Instruct} 18 (x 8 GPUs) hours
\end{itemize}

Our choice of using 8 A100s for \textsc{Llama 3.1 70B Instruct} was due to the resources being available to us, rather than it being necessary for evaluating a model with a beam size of $20$ (in doing so, we expedited our experiments so that we could generate all outputs from Llama 70B on BIRD in under a day). We also note that in many cases, beam search is still a standard (and better performing) strategy for text-to-SQL. In such instances, the only additional inference cost is running nucleus sampling extra $K=10$ times, which in our experiments, takes about 50-70\% less GPU hours compared to running beam search once.

\subsection{Licenses}

Wikidata and Wikipedia are licensed under the Creative Commons CC0 License. \textsc{Llama 3.1} models are licensed under Meta's Llama 3.1 license. The fine-tuned, \textsc{T5 3B PICARD} (\href{https://huggingface.co/tscholak/cxmefzzi}{link}) checkpoint and Hugging Face's transformers library are licensed under Apache 2.0 license. To help parse SQL queries, we use SQLGlot (\href{https://github.com/tobymao/sqlglot}{link}), released under the MIT license. The SPIDER dataset (\href{https://github.com/taoyds/spider/}{link}) is released under Apache 2.0. The BIRD dataset (\href{https://github.com/AlibabaResearch/DAMO-ConvAI/tree/main/bird}{link}) is released under CC BY-SA 4.0.



\end{document}